\documentclass[conference]{IEEEtran}
\IEEEoverridecommandlockouts
\usepackage{cite}
\usepackage{amsmath,amssymb,amsfonts}
\usepackage{algorithmic}
\usepackage{graphicx}
\usepackage{textcomp}
\usepackage{xcolor}
\def\BibTeX{{\rm B\kern-.05em{\sc i\kern-.025em b}\kern-.08em
    T\kern-.1667em\lower.7ex\hbox{E}\kern-.125emX}}

\usepackage{wrapfig}
\usepackage{mathtools}
\usepackage{subcaption}
\usepackage{multirow}
\usepackage{booktabs}

\graphicspath{{figures/}{figures/}}
\def\Agents{\mathcal{I}}
\def\ActionSet{\mathcal{A}}

\usepackage{xspace}

\def\ignore#1{\iffalse #1 \fi\xspace}

\def\ignore#1{\xspace}

\begin{document}

\title{Distributed Multi-Agent Deep Reinforcement Learning for Robust Coordination against Noise}

\author{\IEEEauthorblockN{Yoshinari Motokawa}
\IEEEauthorblockA{\textit{Dept. of Computer Science} \\
\textit{Waseda~University, Tokyo, Japan}\\
y.motokawa@isl.cs.waseda.ac.jp}
\and
\IEEEauthorblockN{Toshiharu Sugawara}
\IEEEauthorblockA{\textit{Dept. of Computer Science} \\
\textit{Waseda~University, Tokyo, Japan}\\
sugawara@waseda.jp}
}


\maketitle

\begin{abstract}
In multi-agent systems, noise reduction techniques are important for improving the overall system reliability as agents are required to rely on limited environmental information to develop cooperative and coordinated behaviors with the surrounding agents. However, previous studies have often applied centralized noise reduction methods to build robust and versatile coordination in noisy multi-agent environments, while distributed and decentralized autonomous agents are more plausible for real-world application. In this paper, we introduce a \emph{distributed attentional actor architecture model for a multi-agent system} (DA3-X), using which we demonstrate that agents with DA3-X can selectively learn the noisy environment and behave cooperatively. We experimentally evaluate the effectiveness of DA3-X by comparing learning methods with and without DA3-X and show that agents with DA3-X can achieve better performance than baseline agents. Furthermore, we visualize heatmaps of \emph{attentional weights} from the DA3-X to analyze how the decision-making process and coordinated behavior are influenced by noise.
\end{abstract}


\section{Introduction}
Noise is unavoidable in real-world applications. In particular, studying noise properties is essential for reinforcement learning. According to Wang et al.~\cite{wang2020reinforcement}, noise in reinforcement learning can be classified into several categories. \emph{Inherent noise} is one of these and represents observational errors resulting from physical conditions such as temperature and lighting. In multi-agent systems, development of noise reduction techniques is crucial for improvement of overall system performance and reliability. Agents are required to rely on limited information to consider the environment and other agents and to develop cooperative and coordinated behavior. Therefore, this technique has been well studied in the \emph{alter-exploration problem}~\cite{matignon_laurent_le_fort-piat_2012}. However, it is difficult to describe such noise properties as its type and frequency of occurrence can only be ascertained from observation.
\par

Analyzing the influence of noise on system performance is an important topic in reinforcement learning, but most research has been conducted from a limited viewpoint. In particular, most previous studies have used the {\em surrogate loss function}~\cite{hendrycks2019using}\ignore{\cite{blanchard2016classification,pmlrv38scott15,hendrycks2019using,vanrooyen2015learning,patrini2017making,zhang2018generalized}} to implement noise reduction. In studies of multi-agent deep reinforcement learning, a few studies~\cite{Ryu_Shin_Park_2020}\ignore{\cite{kilinc2018multiagent,Li_Wu_Cui_Dong_Fang_Russell_2019,Ryu_Shin_Park_2020}} pursue robustness against noisy environment. However, these methods are based on the centralized learner or shared critic networks that provide unbiased feedback to all agents, and require prior knowledge of the domain and noise characteristics to design versatile algorithms. Conversely, distributed/decentralized autonomous agents are more feasible for real-world applications and may make observations with distinct noise levels.
\par

Therefore, we propose a \emph{\textbf{d}istributed \textbf{a}ttentional \textbf{a}ctor \textbf{a}rchitecture model for multi-agent system} (DA3-X) to learn coordinated/cooperative behaviors from noisy information without confusion. The DA3-X is an extension of the \emph{MAT-DQN}~\cite{MotokawaICANN2021} which was proposed to clarify the agent decision-making process using the attention mechanism~\cite{vaswani2017attention}. Unlike the MAT-DQN, any reinforcement learning algorithm such as DA3-DDPG can be applied to our proposed method. We introduce trainableparameters called the {\em saliency vector} embedded in the DA3-X networks to serve as the observation representation during the decision-making process. In our experiments, we assume that agents are unaware how their observation is contaminated by inherent noise or whether surrounding agents are irrelevant or irrational when completing tasks. Thus, agents determine the noise properties in their observations by utilizing the saliency vector to selectively aggregate both reliable and unreliable information.
\par

We then experimentally investigate using the {\em objects collection game} whether agents using our method, {\em DA3-X agents}, can learn appropriate behaviors without confusion, and successfully achieve better performance in noisy environments. The {\em implicit quantile network} (IQN)~\cite{pmlr-v80-dabney18a} and \emph{deep Q-network} (DQN)~\cite{mnih2013atari} were applied to introduce the DA3-IQN and DA3-DQN, and the performance evaluated by comparing them with the vanilla IQN and DQN, respectively. Our experimental results demonstrate that the effect of noise in observation was considerably reduced for agents using DA3-X, and performance exceeded that of the baseline method regardless of reinforcement algorithm. We also analyze the \emph{attentional weights} produced by the DA3-X attention mechanisms for each observed data point. The results indicate that after training higher weighting is assigned to relevant information and lower weighting to unnecessary and/or noisy information. Moreover, by indicating where agents assign attention when making decisions, these results also explain how they are able to act without confusion.
\par

\section{Related Work}
Some research has been done on noise in the field of reinforcement learning~\cite{qlanc,Sabzevari2017ApplicationOR,Minutti2018,NIPS2013_3871bd64}. For example, Natarajan et al.~\cite{NIPS2013_3871bd64} used unbiased estimators and weighted loss functions to investigate how a single agent can be tolerant in the presence of random classification noise by. Additionally, several studies have investigated training with corrupted data and label noise by defining an alternative unbiased surrogate loss function~\cite{hendrycks2019using}. Our goal is to identify methods for agent processing of noisy environments without requiring prior knowledge of noise or changes to loss functions.
\par

Previous work~\cite{plappert2018parameter,fortunato2019noisy,Xu2020ExploringPS} has shown that noise can be helpful by stimulating agents to increase environment exploration. Fortunato et al.~\cite{fortunato2019noisy} proposed the NoisyNet to verify this assertion, adding Gaussian parametric noise to agents' weights and evaluating the improvement in training performance. Han et al.~\cite{han2020nrowandqn} extended the NoisyNet to induce stochasticity in agent policy and develop a noise reduction method for maintaining stability during the training process. However, those methods usually assume that agent observations are always successful, such that if misunderstandings and errors occur in environmental observation by agents, their effective learning fails. Wang et al.~\cite{wang2020reinforcement} developed a robust reinforcement learning framework in which agents are trained using environments with biased noisy rewards. This approach is indeed effective in training with environments such as Atari, but focuses on single agents only.
\par

A few studies on noise in multi-agent systems~\cite{Ryu_Shin_Park_2020} have considered credit assignment or consensus problems for shared centralized networks. In contrast, this paper investigates whether the proposed DA3-X can distinguish between noisy and noiseless data when the individual agent is trained, thereby reducing the possibility of confusion due to noise contamination in the learning process. We also illustrate how DA3-X helped agent estimation of locations of noisy data and other inconsistent agents.
\par

\section{Background and Problem}
\subsection{Dec-POMDP}
The \emph{decentralized partially observable Markov decision process} (dec-POMDP) is an extension of the stochastic Markov decision process (MDP)~\cite{10.5555/528623} wherein the conditional probability distribution of the next step depends only on the current state. Let $t$ ($\geq 0$) be a discrete time-step. The dec-POMDP of $n$ agents, $\Agents = \{1, \ldots, n\}$, is stated by tuple $\langle\Agents, \mathcal{S}, \{\mathcal{A}_i\}, p_T, \{r_i\},\{\Omega_i\}, \mathcal{O}, H\rangle$, where $\mathcal{S}$ is a finite set of available states, $\mathcal{A}_i$ is the set of actions for agent $i\in\Agents$, and $\mathcal{A}=\mathcal{A}_1\times\dots\times\mathcal{A}_n$ being the set of joint actions from policies $\pi=\pi_1\times\cdots\times\pi_n$. Function $p_T(s'|s,a)$ denotes transition probability for $a\in\mathcal{A}$ and $s, s'\in \mathcal{S}$, and $r_i(s, a)$ ($\in\mathbb{R}$) is the reward of $i$ for $s\in\mathcal{S}$ and $a\in\mathcal{A}$. $\Omega=\Omega_1\times\cdots\times\Omega_n$ is the set of joint observations, where $\Omega_i$ is the set of observations for $i$, and $\mathcal{O}(o|s,a)$ is the observation probability $P(o|s,a)$ for $o\in\Omega$. Each agent $i$ aims to maximize the discounted cumulative reward $R_i = \sum_{t=0}^H {\gamma^t r_i(s,a)}$ where $\gamma$ is a discount factor $0\leq \gamma <1$, and $H$ is the process time horizon and also the maximum episode length of a given task. Our objects collection game assumes deterministic behavior, meaning the defined probabilities are set to $0$ or $1$.
\par

\begin{figure}[t]
  \begin{minipage}[t]{0.48\hsize}
    \centering
    \includegraphics[keepaspectratio, width=\linewidth]{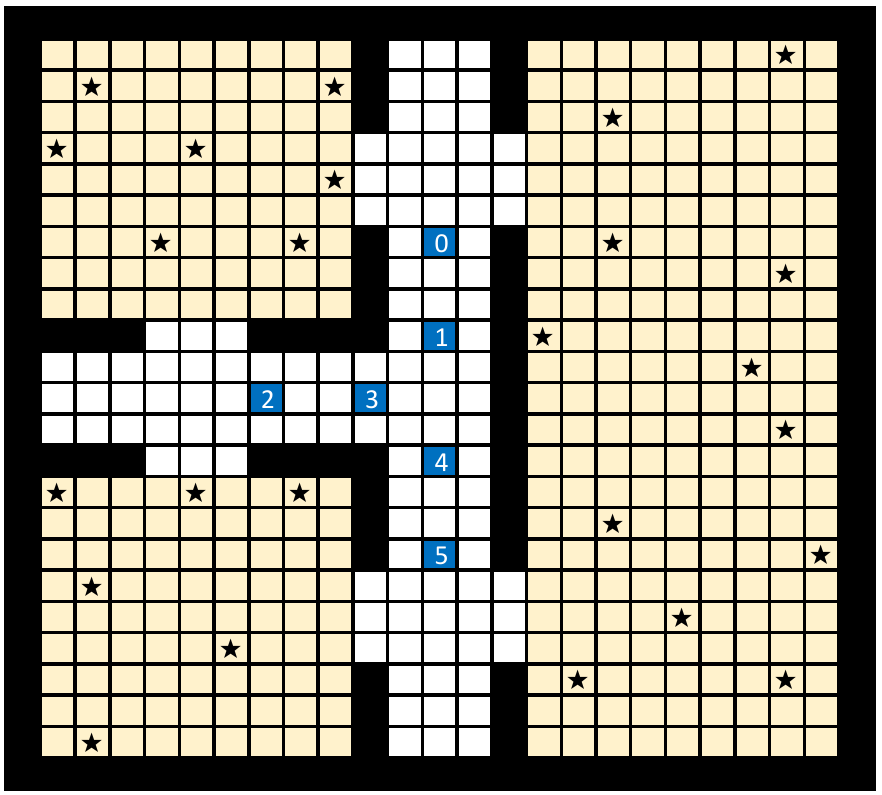}\\
    \subcaption{Three-rooms environment}\label{fig:grid-mapa}
  \end{minipage}
  \hfill
  \begin{minipage}[t]{0.48\hsize}
    \centering
    \includegraphics[keepaspectratio, width=\linewidth]{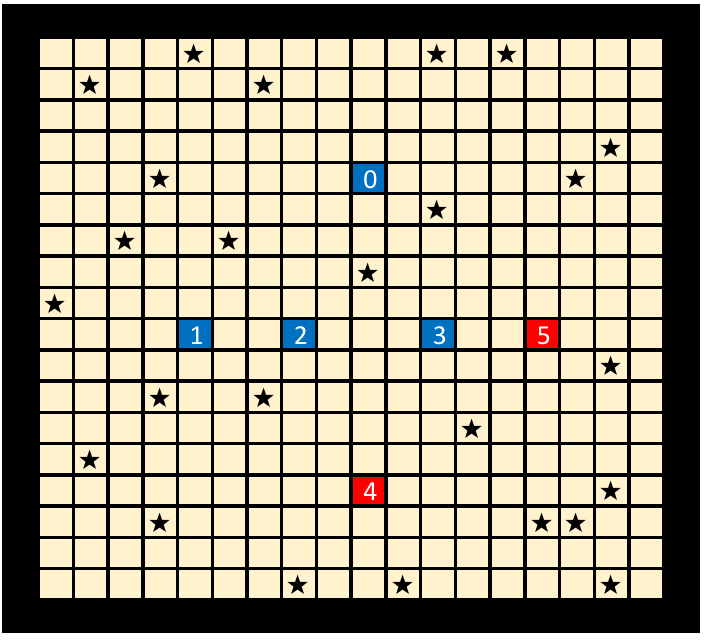}\\
    \subcaption{Simple environment}\label{fig:grid-mapb}
  \end{minipage}
  \begin{minipage}[t]{0.99\hsize}
  \centering
    \includegraphics[keepaspectratio, width=\linewidth]{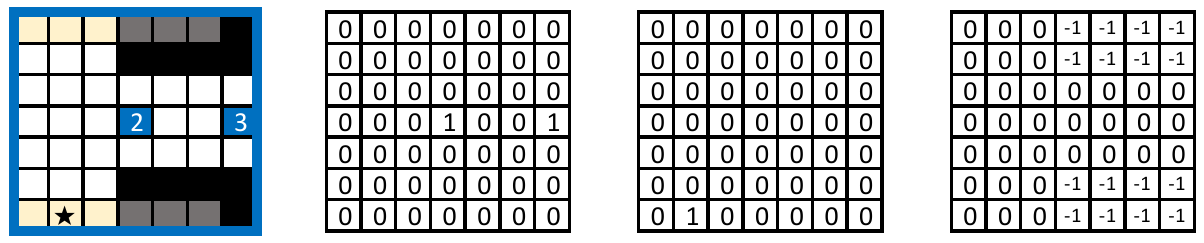}\\
    \subcaption{Encoding local observation when $N_C=3$ and $R_x=R_y=7$}\label{fig:grid-mapc}
  \end{minipage}
    \caption{Grid map environment.}\label{fig:grid-map}
\end{figure}

\subsection{Problem Setting}
In our objects collection game, agents (numbered and colored blue or red) collect scattered objects (described as black stars) placed in a $G_x\times G_y$ grid environment, as illustrated in Fig.~\ref{fig:grid-map}, where $G_x$ and $G_y$ are positive integers. The black, beige, and white nodes in the grid environment describe walls, object areas, and empty areas, respectively. Agent $i$ ($\in\Agents$) can observe the local area within a $R_x\times R_y$ rectangle centered on itself, and is encoded into $N_C$ of $R_x\times R_y$ matrices, which describe the relative positions of other agents, objects, and other components such as walls/obstacles in the visible area. $R_x$ and $R_y$ are positive odd integers and $N_C$ specifies the integer number of matrices. We introduce discrete time whose unit is a time-step, then assume for each time step $i$ takes one of the actions $a_i\in \ActionSet_i = \{\mathit{up}, \mathit{down}, \mathit{right}, \mathit{left}\}$, each of which corresponds to going upward, downward, right, or left. We assume that $i$ picks up an object when $i$ moves to the same node as the object. Then, $i$ receives reward $r_e>0$. However, if $i$ collides with other agents or walls, it receives a small negative reward $r_c$ ($<0$); otherwise, no reward is given. Agents individually utilize their local networks to learn their own policies to effectively collect as many objects as possible and maximize the discounted cumulative reward $R_i$.
\par

\subsection{Attention Mechanisms}
The \emph{self-attention mechanism} draws dependencies between different positions in a single sequence~\cite{vaswani2017attention}. The input sequence of a self-attention network can be categorized into three matrix groups describing the {\em query}, {\em key}, and {\em value}. The compatibility of each query sequence and other components in the key matrix can be obtained by multiplying the query and key matrices. The softmax function is applied to each row of the result to normalize the weights to $[0, 1]$. Subsequently, the normalized weights are multiplied by the logits for determining which information is relevant to agent decision-making processes. The attention is calculated by
\begin{equation}\nonumber
  \mathrm{Attention}(Q,K,V) =
  \mathrm{softmax}(\frac{Q\cdot K^T}{\sqrt{d_k}})V,
\end{equation}
where $Q$, $K$, and $V$ are matrices corresponding to the query, key, and value, respectively, and $d_k$ is the dimension of matrix $K$. Further details are described in ~\cite{vaswani2017attention}.
\par

The \emph{multi-head attention mechanism} is derived by extending the single self-attention function in parallel. Supposing that $h$ attentional heads are available, the multi-heat attention mechanism is expressed by
\begin{equation}
  \begin{split}\nonumber
    \mathrm{MultiHead}(Q, K, V) = \mathrm{Concat}(\mathit{head}_1,
    \ldots, \mathit{head}_h)W^O\\
    \mathit{head}_l = \mathrm{Attention}(Q\cdot W_l^Q, K\cdot W_l^K, V\cdot W_l^V),
  \end{split}
  \label{eq:multi-head-attention}
\end{equation}
where $W_l^Q$, $W_l^K$, $W_l^V$, and $W_l^O$ are projected parameter matrices for $\mathit{head}_l (1 \leq l \leq h)$. As each attentional head characterizes different types of input sequences, agents with the multi-head attention mechanism are expected to flexibly capture the environment by increasing the number of attentional heads $h$, leading to the improved training performance.
\par

\begin{figure}
    \begin{minipage}{0.59\hsize}
      \centering
      \includegraphics[keepaspectratio, width=\linewidth]{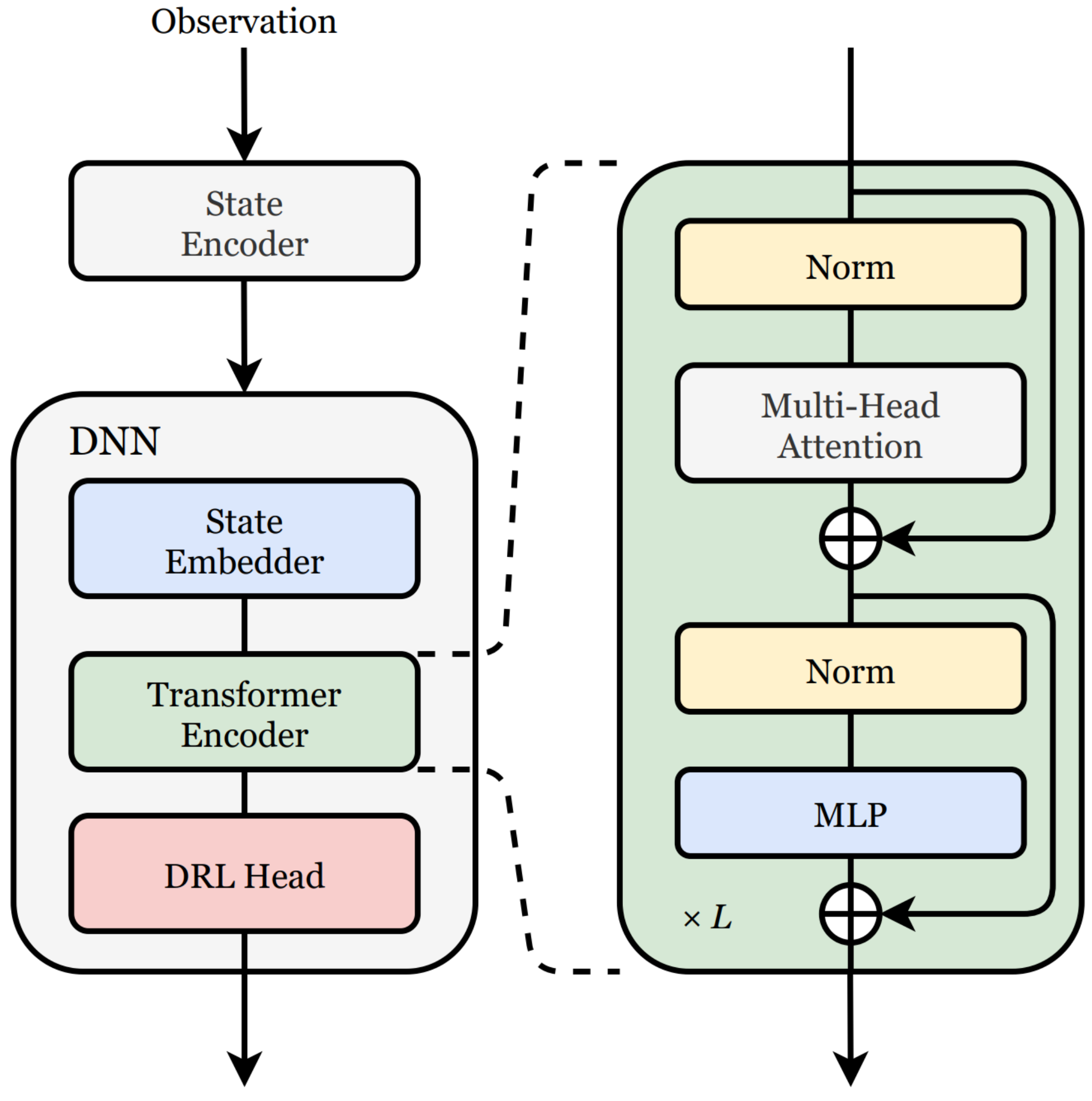}
      \subcaption{DA3-X and transformer encoder}\label{fig:da3_a}
    \end{minipage}
    \hfill
    \begin{minipage}{0.4\hsize}
      \centering
      \includegraphics[keepaspectratio, width=\linewidth]{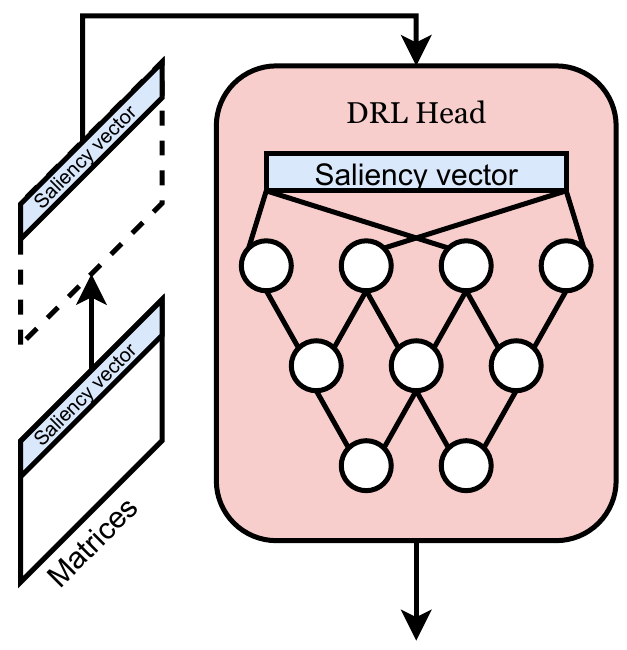}
      \subcaption{Saliency vector fed to DRL head}\label{fig:da3_b}
    \end{minipage}
    \caption{DA3-X structure.}
    \label{fig:da3}
    \end{figure}

\section{Proposed Method}
In the proposed method DA3-X, all agents have their own deep neural network that includes the transformer encoder containing the multi-head attention networks, as illustrated in Fig.~\ref{fig:da3}. Initially, at the \emph{state encoder} the agents encode the observation into $N_C$ of $R_x\times R_y$ matrices that express the agents, objects, and obstacles in the visible local area, as mentioned previously. The encoded matrices comprise a two-valued integer, $\{0, 1\}$ or $\{0, -1\}$, to distinguish the existence of agents, objects, or walls within the visible area. Figure~\ref{fig:grid-mapc} demonstrates the observation encodement by Agent~2 for $N_C=3$ (as illustrated in the leftmost figure in Fig.~\ref{fig:grid-mapc}), where walls are indicated by black nodes, and the invisible area blocked by the wall is indicated by gray in the leftmost figure.
. The first matrix is filled with $1$ at the location of Agent~3, the second with $1$ at the object location, and the third with $-1$ in the region where observation is blocked by walls.
\par

Using a convolutional neural network with kernel size $P$, the encoded $N_C$ matrices are first expanded to $C$ ($>0$) channels of $\lfloor\frac{R_x}{P}\rfloor\times\lfloor\frac{R_y}{P}\rfloor$ matrices, where $P$ is called the \emph{patched size} for the attention mechanism. The matrices are then flattened by the \emph{state embedder}; the matrix $(\lfloor\frac{R_x}{P}\rfloor\cdot\lfloor\frac{R_y}{P}\rfloor + 1) \times C$ is generated. Note that the flattened matrix is multipled by $1$ to introduce the proposed saliency vector into the length of $C$. The saliency vector plays a role similar to that of the \emph{class token} in the BERT~\cite{vaswani2017attention}, and can be implemented similarly to the \emph{classification token} in the vision transformer (ViT)~\cite{dosovitskiy2021image}, but goes further in representing salient pieces of observation (important agents, objects, and walls, etc.), while the BERT class token only represents the overall input text context. Subsequently, the \emph{position embeddings} are added in the matrix like ViT.
\par

The $(\lfloor\frac{R_x}{P}\rfloor\cdot\lfloor\frac{R_y}{P}\rfloor + 1) \times C$ state embedder matrix is fed into the \emph{transformer encoder} as illustrated in Fig.~\ref{fig:da3_a}; $h$ matrices of each \emph{query}, \emph{key}, and \emph{value} whose size is $(\lfloor\frac{R_x}{P}\rfloor\cdot\lfloor\frac{R_y}{P}\rfloor + 1) \times \lfloor \frac{C}{h}\rfloor$ are obtained. The compatibility of each element in local observation, including the saliency vector, is obtained by multiplying the query and key matrices, while the remaining transformer encoder calculations are identical to those in MAT-DQN~\cite{MotokawaICANN2021}. The matrices pass through the transformer encoder $L$ times, where $L$ is a hyper-parameter set depending on the complexity of the learning environment.
\par

Finally, the saliency vector is passed to a deep neural network (DRL) head, as illustrated in Fig.~\ref{fig:da3_b}. Unlike in MAT-DQN, the head can be any network from any deep reinforcement learning algorithm (such as IQN, DDQN, or DDPG) enhancing the learnability and the agent adaptation flexibility. As only the saliency vector represents the transformer encoder output, using the attention mechanism to analyze its mapping with the observation elements allows assessment of the focus of agent attention during decision-making.
\par

\begin{table}
  \caption{Parameter and Value.}\label{table:setting}
  \centering
  \begin{tabular}{lll}
    \toprule
    Description & Parameter & Value \\
    \midrule
    Number of agents & $n$ & 6\\
    Reward for an object collection & $r_e$ & $1$\\
    Reward for collision & $r_c$ & $-1$\\
    Episode length & $H$ & 200 \\
    Epoch & & 5,000\\
    Number of objects & & 25\\
    \bottomrule
  \end{tabular}
\end{table}

\begin{figure}
  \begin{minipage}[t]{0.22\hsize}
    \centering
    \includegraphics[keepaspectratio, width=\linewidth]{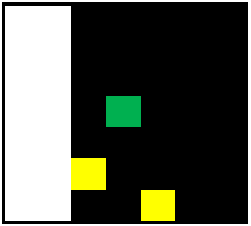}\\
    \subcaption{$R=7$}\label{fig:noises_a}
  \end{minipage}
  \hfill
  \begin{minipage}[t]{0.22\hsize}
    \centering
    \includegraphics[keepaspectratio, width=\linewidth]{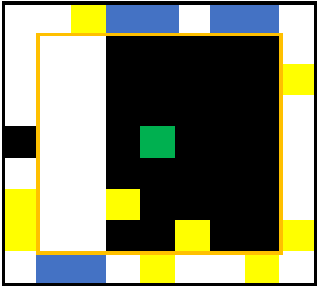}\\
    \subcaption{$R=9$}\label{fig:noises_b}
  \end{minipage}
  \hfill
  \begin{minipage}[t]{0.22\hsize}
    \centering
    \includegraphics[keepaspectratio, width=\linewidth]{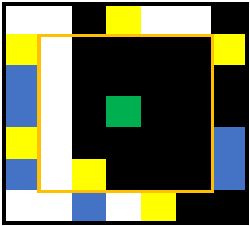}\\
    \subcaption{$R=7$}\label{fig:noises_c}
  \end{minipage}
  \hfill
  \begin{minipage}[t]{0.22\hsize}
    \centering
    \includegraphics[keepaspectratio, width=\linewidth]{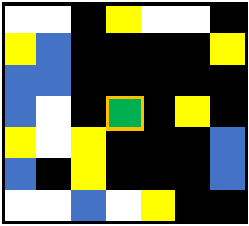}\\
    \subcaption{$R=7$}\label{fig:noises_d}
  \end{minipage}
  \caption{Example noise in observation. (a) Ordinary local observation ($R=7$), where green indicates an observing agent, yellow an object, and white a wall. (b) Observation with large marginal noise ($R=9$). (c) Observation with small marginal noise ($R=7$). (d) Observation with gradual noise ($R=7$).
}
  \label{fig:noises}
\end{figure}

\section{Experiments and Results}
\subsection{Experimental Setup}
Two experiments were conducted using the object collection game in grid-map environments, as illustrated in Fig.~\ref{fig:grid-mapa} and \ref{fig:grid-mapb}, respectively. The purpose of the first experiment (Exp.~1) was to quantitatively evaluate the DA3-X agents when given noisy sensor data, and to identify how they assigned attention in their visible area. In particular, we evaluated DA3-IQN and DA3-DQN as explained later. The second experiment (Exp.~2) aimed to verify whether DA3-X agents could learn coordinated behavior with appropriate agents and ignore irrational agents under the alter-exploration problem. Common parameters in these experiments are listed in Table~\ref{table:setting}. At the beginning of every episode, agents were randomly placed at the numbered nodes in the map (Fig.~\ref{fig:grid-map}), and 25 objects were scattered in the beige region. The agents then started exploration until the time-step $t$ reached episode length $H=200$. When an agent collected an object, $r_e$ was given to the agent and another object was immediately placed in the beige region.
\par

Exp.~1 was conducted in a three-room environment consisting of a $25\times25$ grid (Fig.~\ref{fig:grid-mapa}) to confirm how the agents would share work without becoming confused by noisy data in Env.~1. All six agents ($n=6$) had the same learning capabilities using DA3-IQN, DA3-DQN, or the baseline vanilla DQN and IQN methods. We set $r_c=-1$ and $r_e=1$ as positive and negative rewards, and set $N_C = 3$ matrices, which expressed agents (including themselves at the center), objects, walls, and invisible nodes blocked by walls, as shown in Fig.\ref{fig:grid-mapc}.
\par

Exp.~2 was conducted in the simple $20\times 20$ grid environment (Env.~2) to confirm whether the DA3-X agents could determine which agents were important to coordinate and exclude the non-cooperative irrational agents who could adversely affect the learning process. We set $N_C = n + 2$ to distinguish other agents, i.e., two matrices for objects and invisible nodes as in the Exp.~1, and $n$ matrices to observe their own location and those of other nearby agents. Accordingly, we assumed that agents had IDs, and could identify other agent IDs in the visible area. We also assumed that two of the six agents would wander without performing work; thus, those agents are considered not worth including in coordination.
\par

The DA3-X agents were given four attentional heads ($h=4$), and the loop count through the transformer encoder was set to $L=1$. For the baseline methods, the vanilla IQN and DQN structures asides from their DRL heads were identical to those in Fig.~\ref{fig:da3}, but did not feature the transformer encoder component. The IQN head for DA3-IQN was built using the DDQN~\cite{hasselt2015doubledqn} and dueling network~\cite{dueling} algorithm. DA3-DQN utilized only a fully connected network for the DRL head, while the vanilla DQN comprised two convolutional layers, two max-pooling layers, and a fully connected network without the dueling network algorithm. The chosen patched size was $P=1$ so that all elements in the observation were associated with the attention mechanism. The length of the saliency vector was set to $C=64$, and the visible area comprised a square of side length $R$ (=$R_x=R_y$).
\par

\begin{figure*}
\begin{minipage}[t]{0.5\hsize}
    \centering
    \includegraphics[keepaspectratio, width=1.1\linewidth]{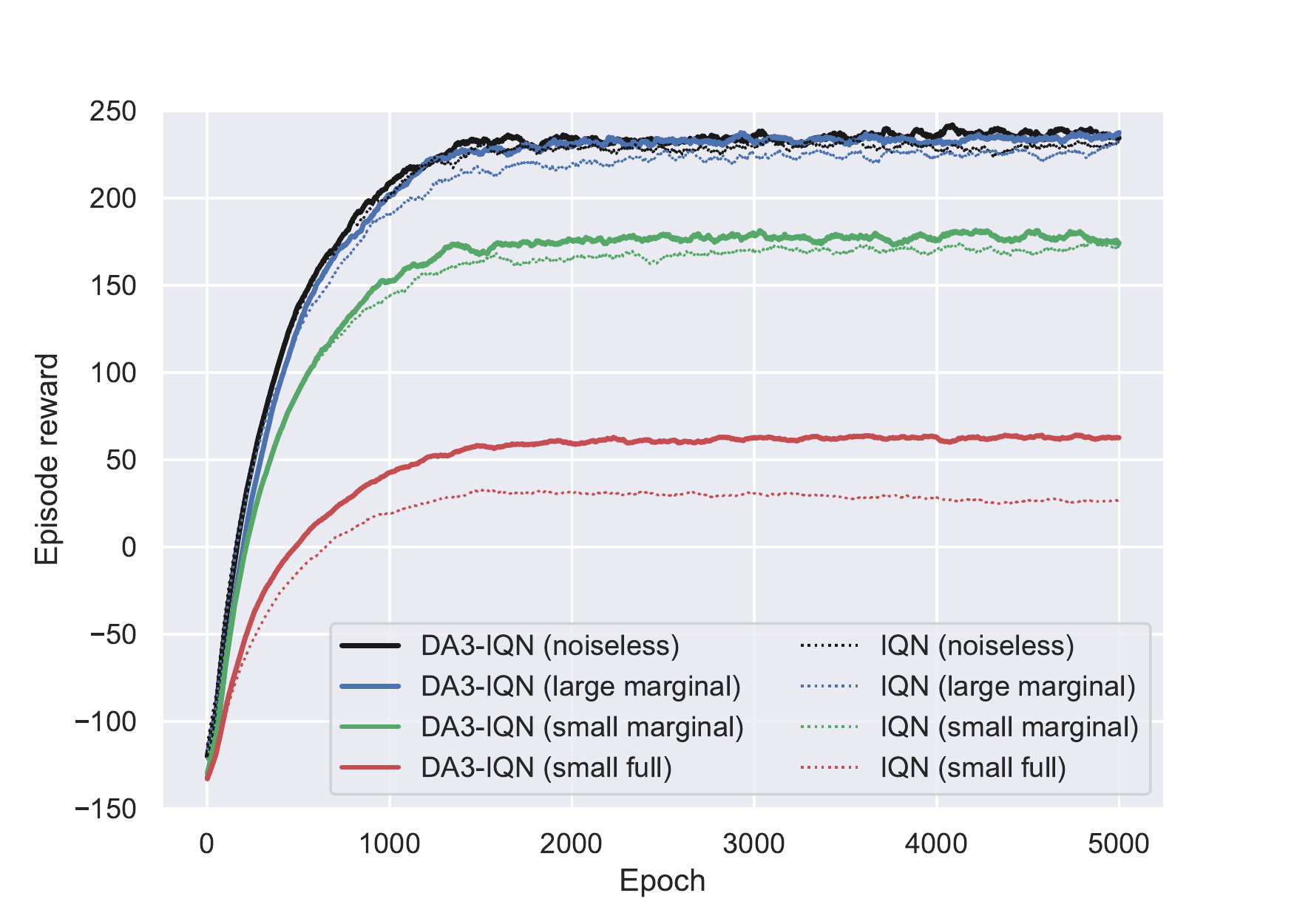}\\
    \subcaption{Episode reward by IQN head}
\end{minipage}
\hfill
\begin{minipage}[t]{0.5\hsize}
    \centering
    \includegraphics[keepaspectratio, width=1.1\linewidth]{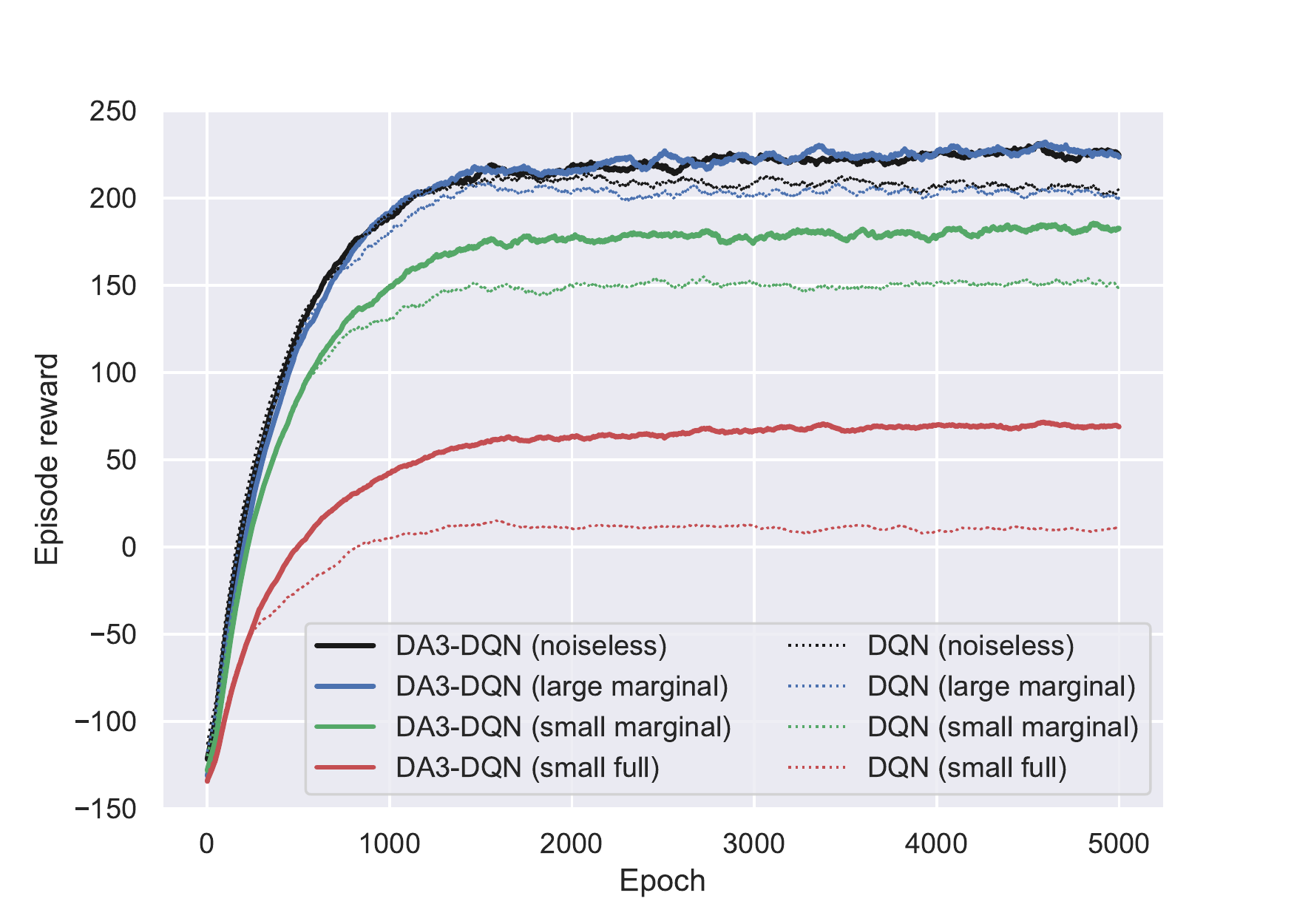}\\
    \subcaption{Episode reward by DQN head}
\end{minipage}
\caption{Learning performance with varying observation noise.}
\label{fig:performance_comparison}
\end{figure*}

\subsection{Noise Definition}
\subsubsection{Noise from Sensors}
In Exp.~1, agents observe an environment with flip-noise inspired by \cite{NIPS2013_3871bd64}; the existence of agents, objects, or walls are stochastically flipped between 0 and 1, or between 0 and -1 in encoded matrices. Therefore, agents are subject to the uncertainty of whether a visible entity is genuine or noise. Figure~\ref{fig:noises} illustrates the effect of noise on observations. Figure~\ref{fig:noises_a} is a snapshot of local observation (without noise) of a particular agent when $R=7$, where the central green node is the agent, blue nodes are other agents, yellow nodes are objects to be collected, black nodes are empty (no data), and white nodes are walls.
\par

The \emph{marginal noise} model is introduced, in which sensor sensitivity decreases with distance, with especially higher noise around sensible region boundaries. Fig.~\ref{fig:noises_b} ($R=9$) shows that the data in the peripheral range (outside the orange square) have high marginal noise and so are flipped with probability 0.5, implying that the data in this area are random and contain no information. This is called the {\em large marginal noise observation}. Note that this observation contains the same information quantity as the {\em noiseless observation} with $R=7$ shown in Fig.~\ref{fig:noises_a}. Fig.~\ref{fig:noises_c} ($R=7$) is the {\em small marginal noise observation} that features low peripheral range noise with a probability of flipping of 0.2. We also consider the \emph{small full noise observation} in which observed data outside the orange square are flipped with stepwise probabilities as shown in Fig.~\ref{fig:noises_d}, i.e., the probability of flipping is 0.2, 0.1 and 0.05 for the outermost, middle and inner regions respectively.
\par

\subsubsection{Noisy Agents}
Exp.~2 considers noise in terms of reliability for coordination building. In multi-agent systems, agent trust is required for cooperative behavior. As described in the alter-exploration problem, if a co-working agent is broken or takes unexpected and/or unintelligent actions, the learning of cooperative behavior is impaired and doomed to failure. Therefore, we introduced two wandering agents that do not collect objects to investigate how the DA3-X agents can mitigate this issue.
\par

\begin{table}[]
\caption{Quantitative performance with noise in detail.}\label{table:result-exp1}
\subcaption{Performance by DA3-IQN}\label{table:result-exp1-DA3-IQN}
\resizebox{\linewidth}{!}{\begin{tabular}{llll}
    \toprule
    Noise & Objects collected & Agents collision & Walls collision \\
    \midrule
    Noiseless  & $\mathbf{239.86\pm 21.46}$    & $1.13\pm 0.39$    & $3.30\pm 1.32$     \\
    Large marginal & $240.23\pm 20.33$    & $1.20\pm 0.39$    & $3.10\pm 0.59$      \\
    Small marginal & $178.90\pm 18.64$    & $1.05\pm 0.34$    & $2.71\pm 0.79$    \\
    Small full     & $\mathbf{75.55\pm 6.65}$     & $\mathbf{7.33\pm 1.33}$     & $\mathbf{5.34\pm 0.77}$    \\ \bottomrule
\end{tabular}}\\[5pt]
\subcaption{Performance by IQN (baseline)}\label{table:result-exp1-Vanilla-IQN}
\resizebox{\linewidth}{!}{%
    \begin{tabular}{llll}
    \toprule
    Noise   & Objects collected & Agents collision & Walls collision \\
    \midrule
    Noiseless    & $235.09\pm 19.33$    & $0.91\pm 0.32$    & $3.43\pm 0.86$ \\
    Large marginal & $234.84\pm 18.02$    & $1.16\pm 0.39$    & $3.57\pm 0.68$ \\
    Small marginal & $176.48\pm 14.84$    & $1.18\pm 0.36$    & $2.99\pm 0.75$ \\
    Small full& $47.71\pm 4.02$ & $11.81\pm 2.66$ & $9.62\pm 1.33$ \\
    \bottomrule
\end{tabular}}\\[5pt]
\subcaption{Performance by DA3-DQN}\label{table:result-exp1-DA3-DQN}
\resizebox{\linewidth}{!}{%
    \begin{tabular}{llll}
    \toprule
    Noise   & Objects collected & Agents collision & Walls collision \\
    \midrule
    Noiseless    & $\mathbf{232.12\pm 17.70}$    & $1.22\pm 0.40$    & $4.11\pm 2.18$ \\
    Large marginal & $230.37\pm 18.37$    & $1.28\pm 0.33$    & $2.77\pm 0.69$ \\
    Small marginal & $185.19\pm 16.65$    & $0.96\pm 0.32$    & $2.57\pm 0.74$ \\
    Small full& $\mathbf{81.01\pm 6.24}$ & $\mathbf{7.16\pm 1.49}$ & $\mathbf{4.76\pm 0.72}$ \\
    \bottomrule
\end{tabular}}\\[5pt]
\subcaption{Performance by DQN (baseline)}\label{table:result-exp1-Vanilla-DQN}
\resizebox{\linewidth}{!}{%
    \begin{tabular}{llll}
    \toprule
    Noise   & Objects collected & Agents collision & Walls collision \\
    \midrule
    Noiseless    & $209.77\pm 20.63$    & $1.62\pm 0.42$    & $4.63\pm 1.76$ \\
    Large marginal & $208.02\pm 18.01$    & $1.89\pm 0.51$    & $5.01\pm 1.08$ \\
    Small marginal & $156.85\pm 14.54$    & $1.69\pm 0.46$    & $3.75\pm 0.69$ \\
    Small full& $41.88\pm 3.94$ & $18.16\pm 3.45$ & $13.28\pm 1.45$ \\
    \bottomrule
\end{tabular}}
\end{table}

\subsection{Performance Comparison -- Exp.~1}
We investigated whether the DA3-IQN and DA3-DQN agents could learn over time using noiseless data and three types of noisy data shown in Fig.~\ref{fig:noises}a-d. The data presented below are the averaged values for 10 simulation trials.
\par

Figure~\ref{fig:performance_comparison} plots the earned rewards per episode for the DA3-IQN, DA3-DQN, vanilla IQN, and vanilla DQN agents in our environments. Table~\ref{table:result-exp1} lists the averaged numbers of collected objects and collisions with agents/walls with their standard deviations for the final 100 episodes. In general, the learning performance in the object collection game is expected to decrease when the noise in agent observations is more likely , regardless of reinforcement learning method, and we have observed this phenomenon in Exp.~1.
\par

Figure~\ref{fig:performance_comparison} and Table~\ref{table:result-exp1} indicate that the DA3-IQN and DA3-DQN agents achieved higher episode rewards than the baseline agents in all environments, implying more efficient object collection with fewer collisions. As presented in Table~\ref{table:result-exp1}, the differences in the number of collected objects by DA3-IQN versus IQN agents are $4.77$ (noiseless, improvement of $2.03\%$), $5.39$ (large marginal noise, $2.30\%$), $2.42$ (small marginal noise, $1.37\%$), and $27.84$ (small full noise, improvement of $58.35\%$). For DA3-DQN versus DQN agents, the differences are $22.36$ (noiseless, improvement of $10.66\%$), $22.35$ (large marginal noise, $10.74\%$), $28.34$(small marginal noise, $18.07\%$), and $39.13$ (small full noise, improvement of $93.44\%$). The DA3-IQN and DA3-DQN agents also achieved fewer collisions than the baseline agents, especially in the small full noise environments.
\par

The use of DA3-X clearly mitigates the negative effect of noise in observation, as the performance improvement becomes more pronounced as the influence of noise increases. The DA3-X agents are likely coordinating by utilizing the saliency vector in the attention mechanism to identify the necessary information components in their observation to induce better policies. Therefore, DA3-X agents can deliberately limit the observation component that they examine closely for their decision-making (or conversely ignore), resulting in faster training and effective behavior with fewer collisions with other agents or walls even in very noisy environments. This feature will be discussed further in Section~\ref{section:analysis_attention}.
\par

The results show DA3-X improved the learning performance even for noiseless observations. In particular, the performance difference in number of collected objects between DA3-IQN and DA3-DQN ($7.74$) became smaller than that between IQN and DQN ($25.32$). It is also notable that DA3-DQN agents could perform as well as the DA3-IQN or vanilla IQN agents, despite the learnability of IQN previously reported~\cite{pmlr-v80-dabney18a} as higher than that of DQN and the DA3-DQN agents using simple DQN network as the DRL head. Therefore, these results suggest that DA3-X enhanced the learning performance of weak learners such as DQN.
\par

\begin{figure}
\begin{minipage}[t]{0.4\hsize}
    \centering
    \includegraphics[keepaspectratio, width=0.8\linewidth]{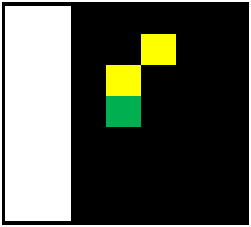}\\
    \subcaption{Observation}\label{fig:none_hm_a}
\end{minipage}
\begin{minipage}[t]{0.50\hsize}
    \centering
    \includegraphics[keepaspectratio, width=\linewidth]{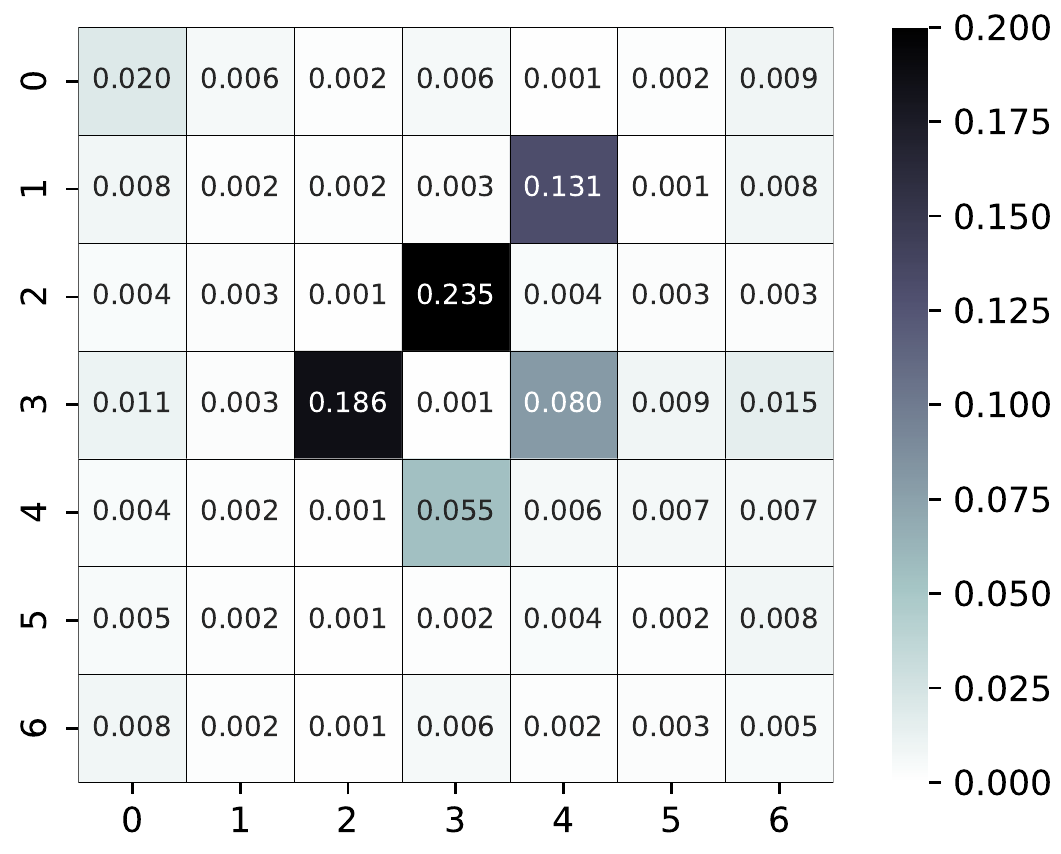}\\
    \subcaption{Agent's attention}\label{fig:none_hm_b}
\end{minipage}
\caption{Attentional heatmap (noiseless).}
\label{fig:none_hm}
\end{figure}

\begin{figure}
\begin{minipage}[t]{0.4\hsize}
    \centering
    \includegraphics[keepaspectratio, width=0.8\linewidth]{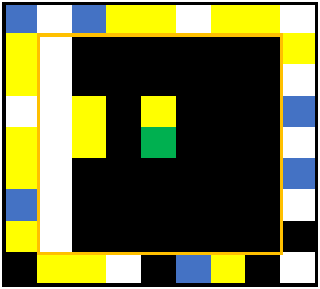}\\
    \subcaption{Observation}\label{fig:9_0.5_hm_a}
\end{minipage}
\begin{minipage}[t]{0.5\hsize}
    \centering
    \includegraphics[keepaspectratio, width=\linewidth]{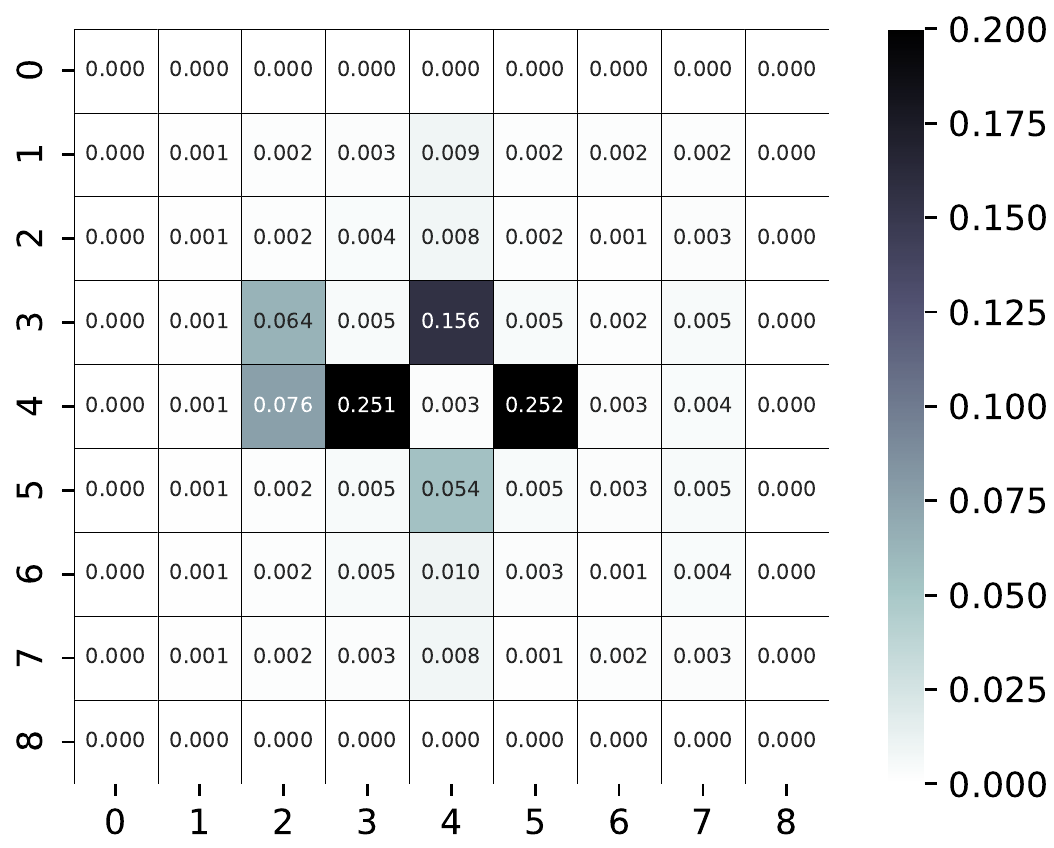}\\
    \subcaption{Agent's attention}\label{fig:9_0.5_hm_b}
\end{minipage}
\caption{Attentional heatmap (large marginal noise).}
\label{fig:9_0.5_hm}
\end{figure}

\subsection{Analysis of Attention}\label{section:analysis_attention}
DA3-DQN agents were able to achieve large performance improvements, particularly for the small marginal and small full noise observation cases. Therefore we analyzed how these agents handled a contaminated observation during their decision-making process. Accordingly, attentional heatmaps were generated from the DA3-DQN transformer encoder and visualized. Figure~\ref{fig:none_hm} illustrates (a) the local observation of a particular agent and (b) the heatmap of the mean attentional weights from four attentional heads ($h=4$) in a noiseless environment.
\par

An example of a noiseless environment observation is illustrated in Fig.~\ref{fig:none_hm_a}. It indicates that the agent observes the walls on the left side and two objects that are located above. Fig.~\ref{fig:none_hm_b} is the attentional heatmap indicating the regions of agent focus in their visible area for the same situation. The heatmap shows that the agent assigns relatively high attention to (1) its four neighbor nodes, one of which being the next node to move, and (2) object locations; thus, the upper node attentional weight is highest and the agent moves up to earn the immediate $r_e$ for the next action.
\par

Figure~\ref{fig:9_0.5_hm} shows the DA3-DQN agent areas of focus during observation for the large marginal noise environment. It is evident that the agent similarly focuses attention to its four neighbor nodes and the nodes containing objects. Remarkably, almost all attentional weights in the outermost region of the visible area are $0.000$, meaning that the agent recognizes that no useful information is located there and thus learns to ignore them. This fact explains the result that DA3-IQN and DA3-DQN agents exhibited almost identical performances in both the noiseless environment with $R=7$ and the large marginal noise environment with $R=9$ (Table~\ref{table:result-exp1}).
\par

\begin{figure}
\begin{minipage}[t]{0.4\hsize}
    \centering
    \includegraphics[keepaspectratio, width=0.8\linewidth]{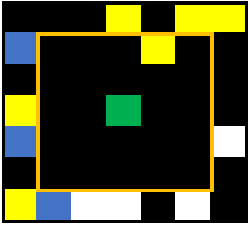}\\
    \subcaption{Observation}\label{fig:7_0.2_hm_a}
\end{minipage}
\begin{minipage}[t]{0.50\hsize}
    \centering
    \includegraphics[keepaspectratio, width=\linewidth]{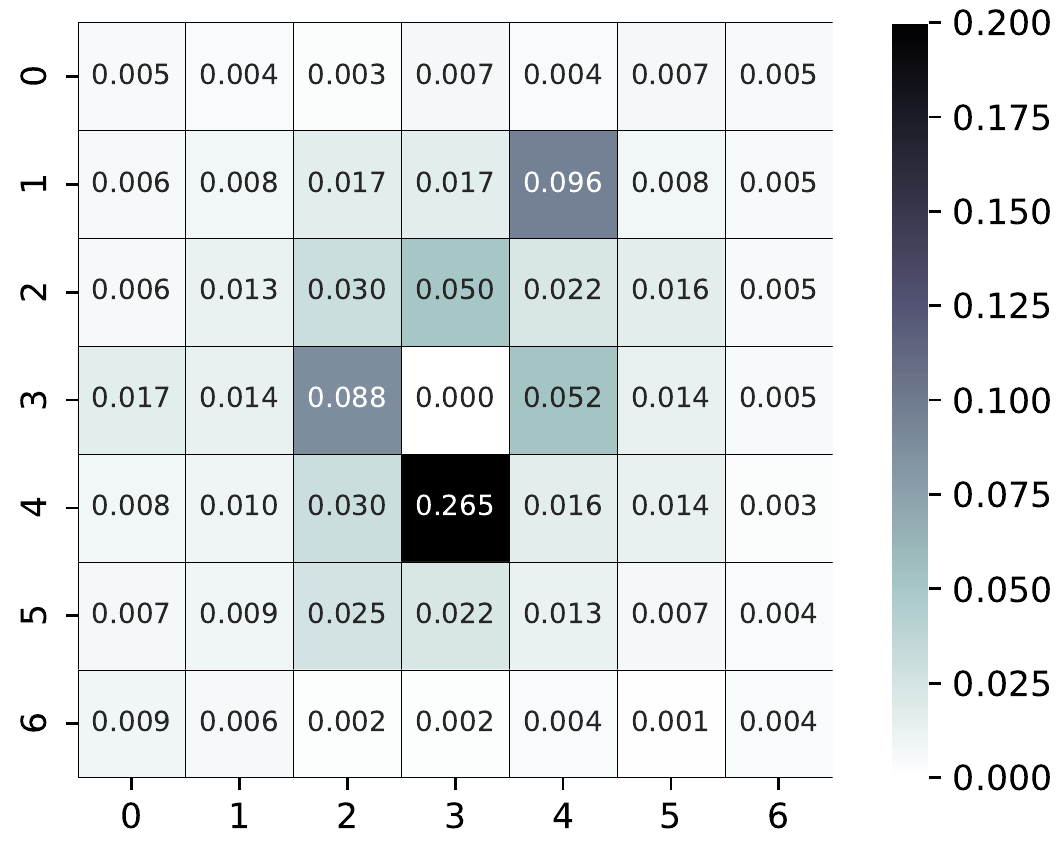}\\
    \subcaption{Agent's attention}\label{fig:7_0.2_hm_b}
\end{minipage}
\caption{Attentional heatmap (small marginal noise).}
\label{fig:7_0.2_hm}
\end{figure}

\begin{figure}
\begin{minipage}[t]{0.4\hsize}
\centering
\includegraphics[keepaspectratio, width=0.8\linewidth]{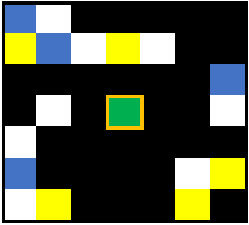}\\
\subcaption{Observation}\label{fig:7_0.2s_hm_a}
\end{minipage}
\begin{minipage}[t]{0.50\hsize}
\centering
\includegraphics[keepaspectratio, width=\linewidth]{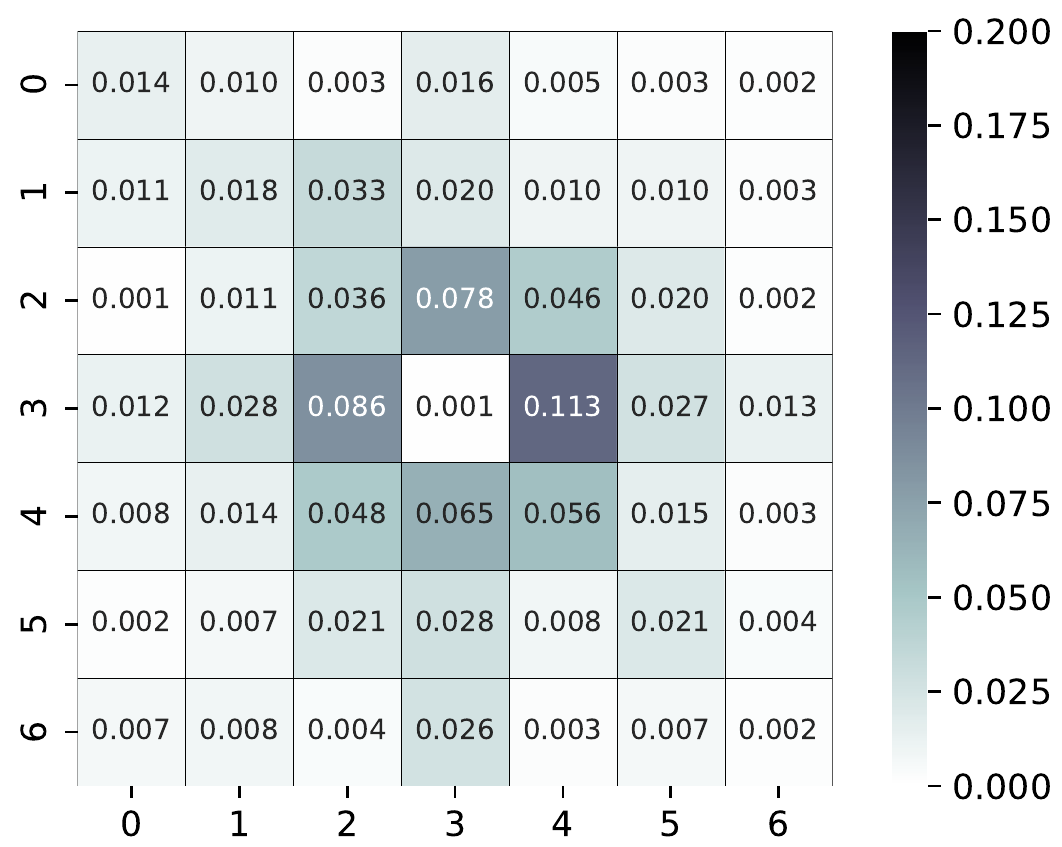}\\
\subcaption{Agent's attention}\label{fig:7_0.2s_hm_b}
\end{minipage}
\caption{Attentional heatmap (small full noise).}
\label{fig:7_0.2s_hm}
\end{figure}

The DA3-DQN agent method of dealing with small marginal noise seems consistent, as shown in Fig.~\ref{fig:7_0.2_hm}. The agent again assigns relatively high attention to its neighbor nodes ($0.052$, $0.050$, $0.088$ and $0.265$) and to object locations, as for the large marginal noise environment. Interestingly, elsewhere the agent assigns overall attentional weights in the range of $[0.001, 0.030]$, slightly higher than those in the first and second environments (see Figs.~\ref{fig:none_hm_b} and \ref{fig:9_0.5_hm_b}), which is likely due to small marginal noise in the outermost region. This suggests that the agent's attention has been partially distracted, but is nevertheless able to set attentional weights appropriately as the area excluding the outermost area contains no noise.
\par

Finally, when observing an environment with low full noise (Fig.~\ref{fig:7_0.2s_hm_a}),the trained DA3-DQN agents focused on eight surrounding nodes equally, as shown in Fig.~\ref{fig:7_0.2s_hm_b}. The mean attentional weights in the outermost region were approximately $0.006$, while those in the middle and inner regions were approximately $0.018$ and $0.066$, respectively. The agents successfully chose their actions with full noise, focusing mostly on the relatively reliable region its surrounding eight nodes, and then on the middle region with lower attentional weights. This was quite different from the other environments in which agents gave high weights to four adjacent nodes as the next nodes to move. Relatively larger weights were not given to the nodes where agents/objects exist. Nevertheless, the performance of the DA3-DQN agents was much better than that of the baseline agents, with the DA3-DQN agents collecting over $81$ objects, whereas the baseline agents collected fewer than $42$ objects.
\par

Thus, DA3-X agents successfully built policies by excluding or slightly referring to noisy regions of observations. We verified that DA3-X agents could use noise to infer the uncertainty of information through thousands of experiences, which led to the attentional weights focusing on higher or more confident regions of observations.
\par

\begin{figure}[t]
\begin{minipage}[t]{0.48\hsize}
    \centering
    \begin{minipage}[t]{0.35\hsize}
    \centering
    \includegraphics[keepaspectratio, width=\linewidth]{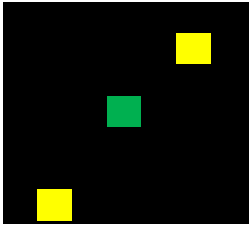}
    \end{minipage}
    \hfil\hfil
    \begin{minipage}[t]{0.35\hsize}
    \centering
    \includegraphics[keepaspectratio, width=\linewidth]{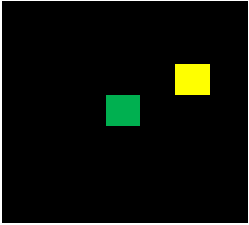}
    \end{minipage}\hfil\hfil \\
    \begin{minipage}[t]{0.48\hsize}
    \centering
    \includegraphics[keepaspectratio, width=\linewidth]{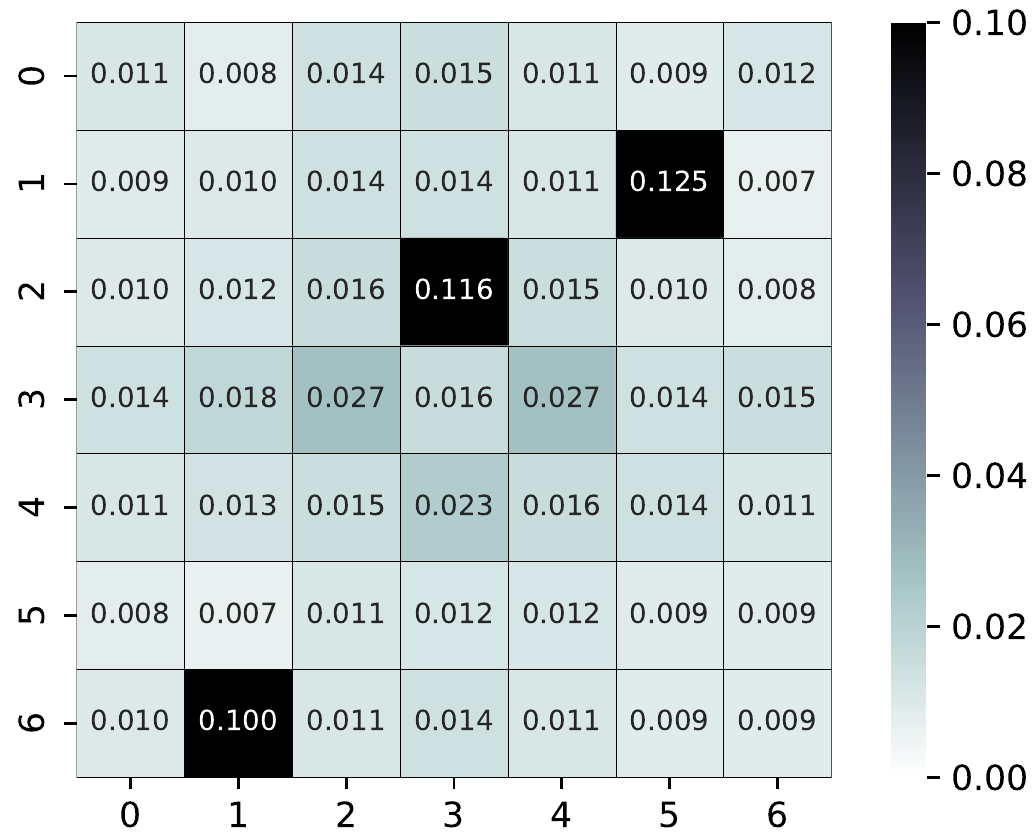}
    \end{minipage}
    \begin{minipage}[t]{0.48\hsize}
    \centering
    \includegraphics[keepaspectratio, width=\linewidth]{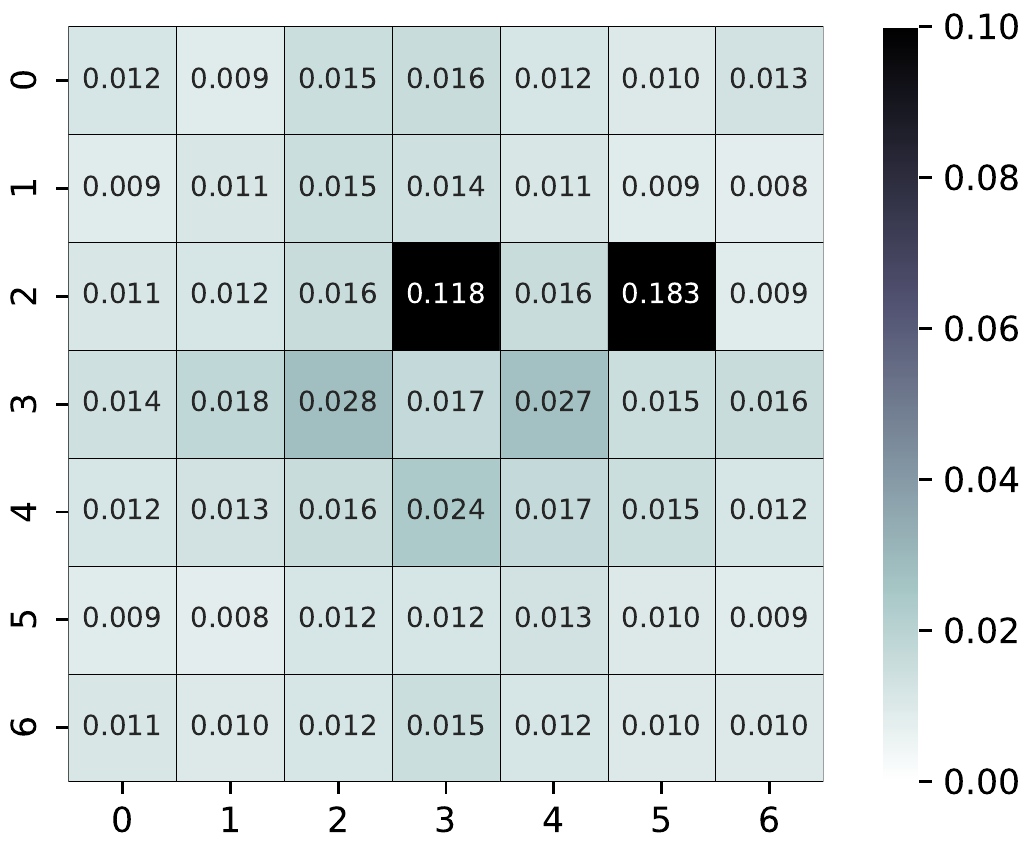}
    \end{minipage}\\
    \subcaption{Situation~1: Move up}\label{fig:coordination_test_a}
\end{minipage}
\hfill
\begin{minipage}[t]{0.48\hsize}
    \centering
    \begin{minipage}[t]{0.35\hsize}
    \centering
    \includegraphics[keepaspectratio, width=\linewidth]{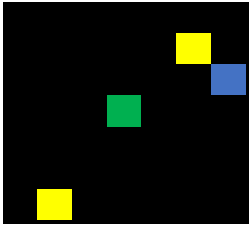}
    \end{minipage}
    \hfil\hfil
    \begin{minipage}[t]{0.35\hsize}
    \centering
    \includegraphics[keepaspectratio,
      width=\linewidth]{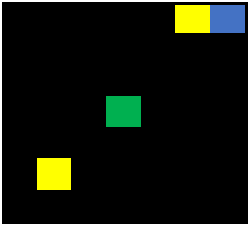}
    \end{minipage}\hfil \\
    \begin{minipage}[t]{0.48\hsize}
    \centering
    \includegraphics[keepaspectratio, width=\linewidth]{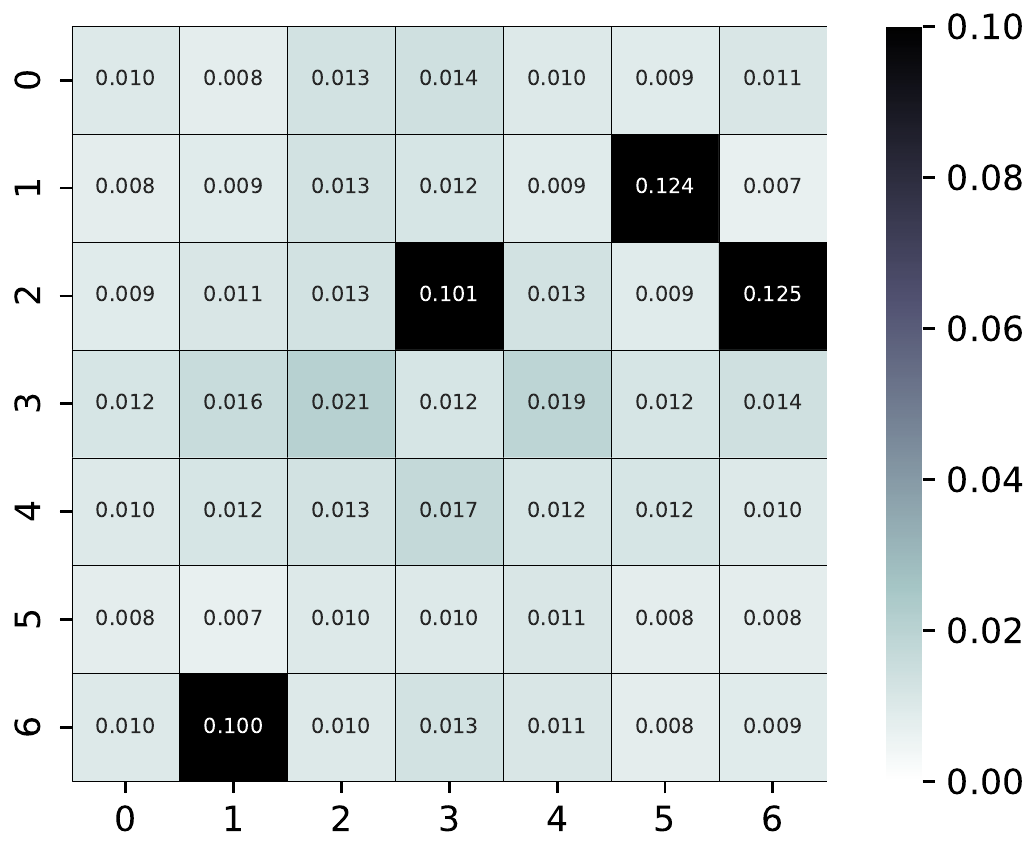}
    \end{minipage}
    \begin{minipage}[t]{0.48\hsize}
    \centering
    \includegraphics[keepaspectratio, width=\linewidth]{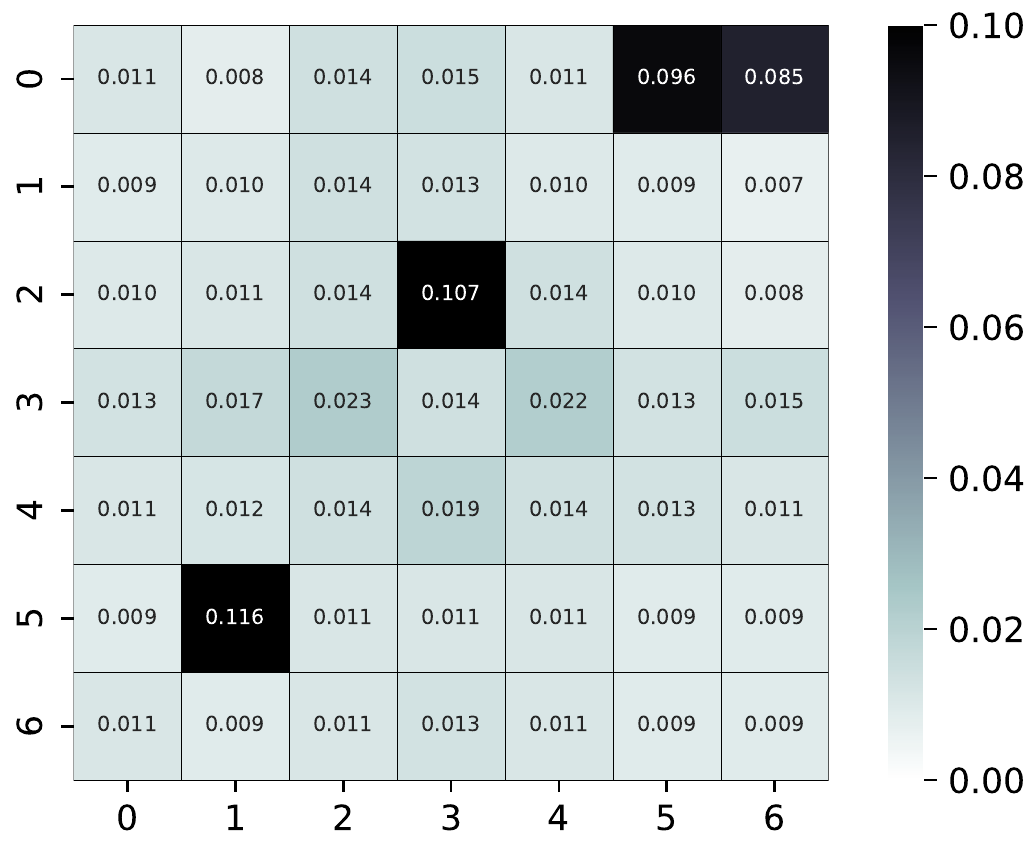}
    \end{minipage}\\
    \subcaption{Situation~2: Move down}\label{fig:coordination_test_b}
\end{minipage}\\[5pt]
\begin{minipage}[t]{0.48\hsize}
    \centering
    \begin{minipage}[t]{0.35\hsize}
    \centering
    \includegraphics[keepaspectratio, width=\linewidth]{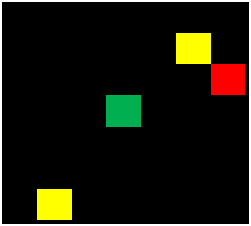}
    \end{minipage}
    \hfil\hfil\hfil
    \begin{minipage}[t]{0.35\hsize}
    \centering
    \includegraphics[keepaspectratio, width=\linewidth]{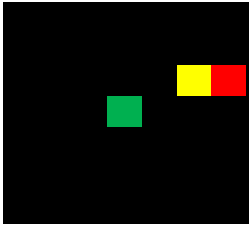}
    \end{minipage}\hfil \\
    \begin{minipage}[t]{0.48\hsize}
    \centering
    \includegraphics[keepaspectratio, width=\linewidth]{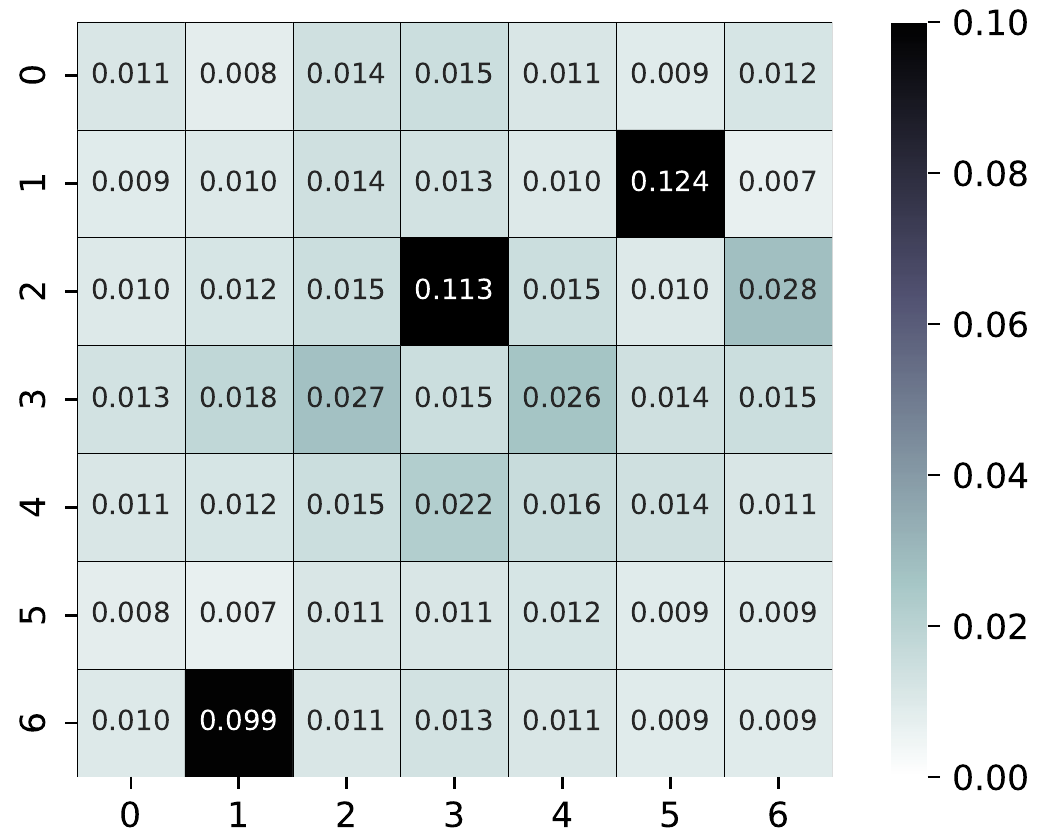}
    \end{minipage}
    \begin{minipage}[t]{0.48\hsize}
    \centering
    \includegraphics[keepaspectratio, width=\linewidth]{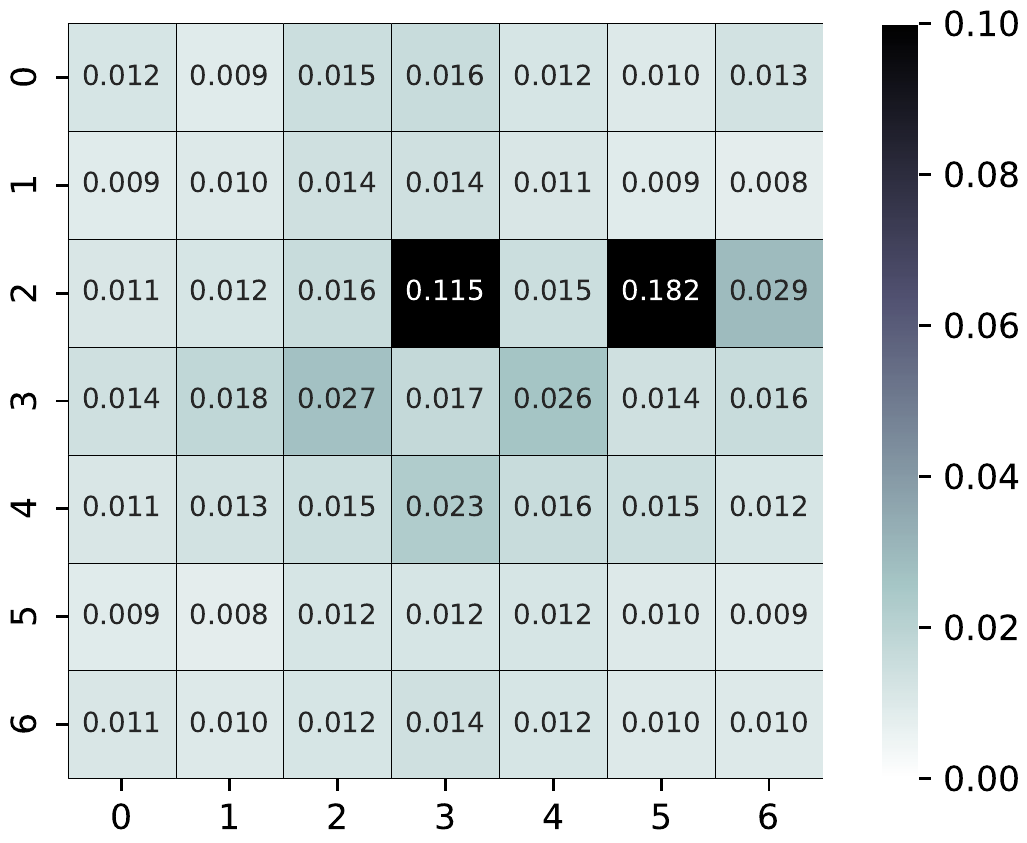}
    \end{minipage}\\
    \subcaption{Situation~3: Move up}\label{fig:coordination_test_c}
\end{minipage}
\hfill
\begin{minipage}[t]{0.48\hsize}
    \centering
    \begin{minipage}[t]{0.35\hsize}
    \centering
    \includegraphics[keepaspectratio, width=\linewidth]{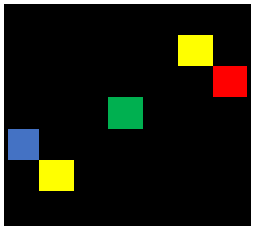}
    \end{minipage}
    \hfil\hfil\hfil
    \begin{minipage}[t]{0.35\hsize}
    \centering
    \includegraphics[keepaspectratio, width=\linewidth]{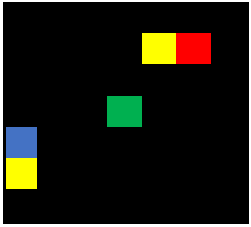}
    \end{minipage}\hfil \\
    \begin{minipage}[t]{0.48\hsize}
    \centering
    \includegraphics[keepaspectratio, width=\linewidth]{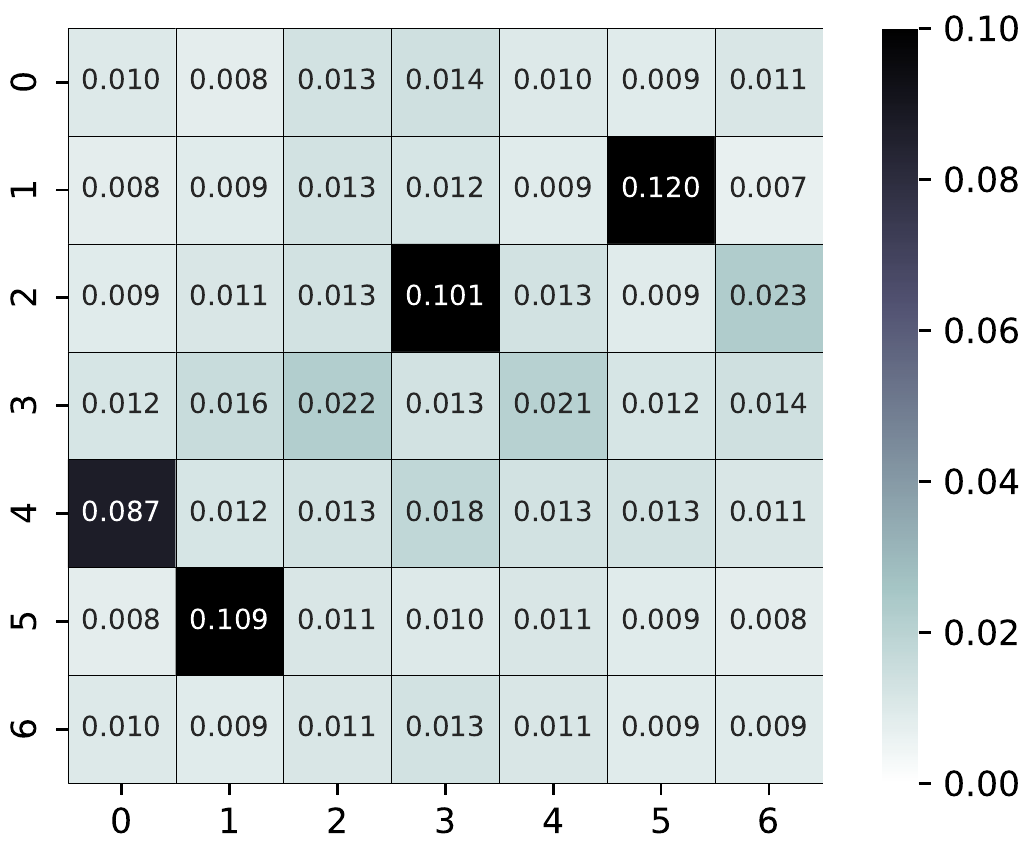}
    \end{minipage}
    \begin{minipage}[t]{0.48\hsize}
    \centering
    \includegraphics[keepaspectratio, width=\linewidth]{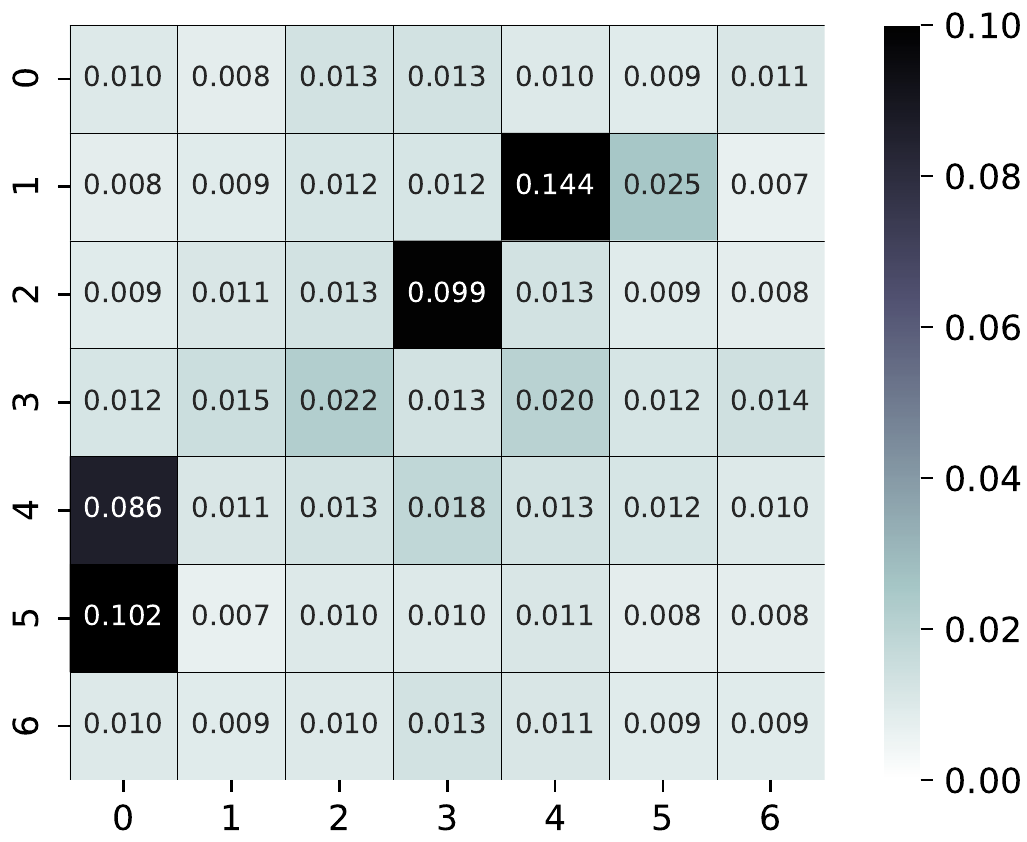}
    \end{minipage}\\
    \subcaption{Situation~4: Move right}\label{fig:coordination_test_d}
\end{minipage}
\caption{Attentional heatmaps and agent behaviors.}
\label{fig:coordination_test}
\end{figure}

\begin{figure}[t]
\begin{minipage}[t]{0.31\hsize}
    \centering
    \includegraphics[keepaspectratio, width=\linewidth]{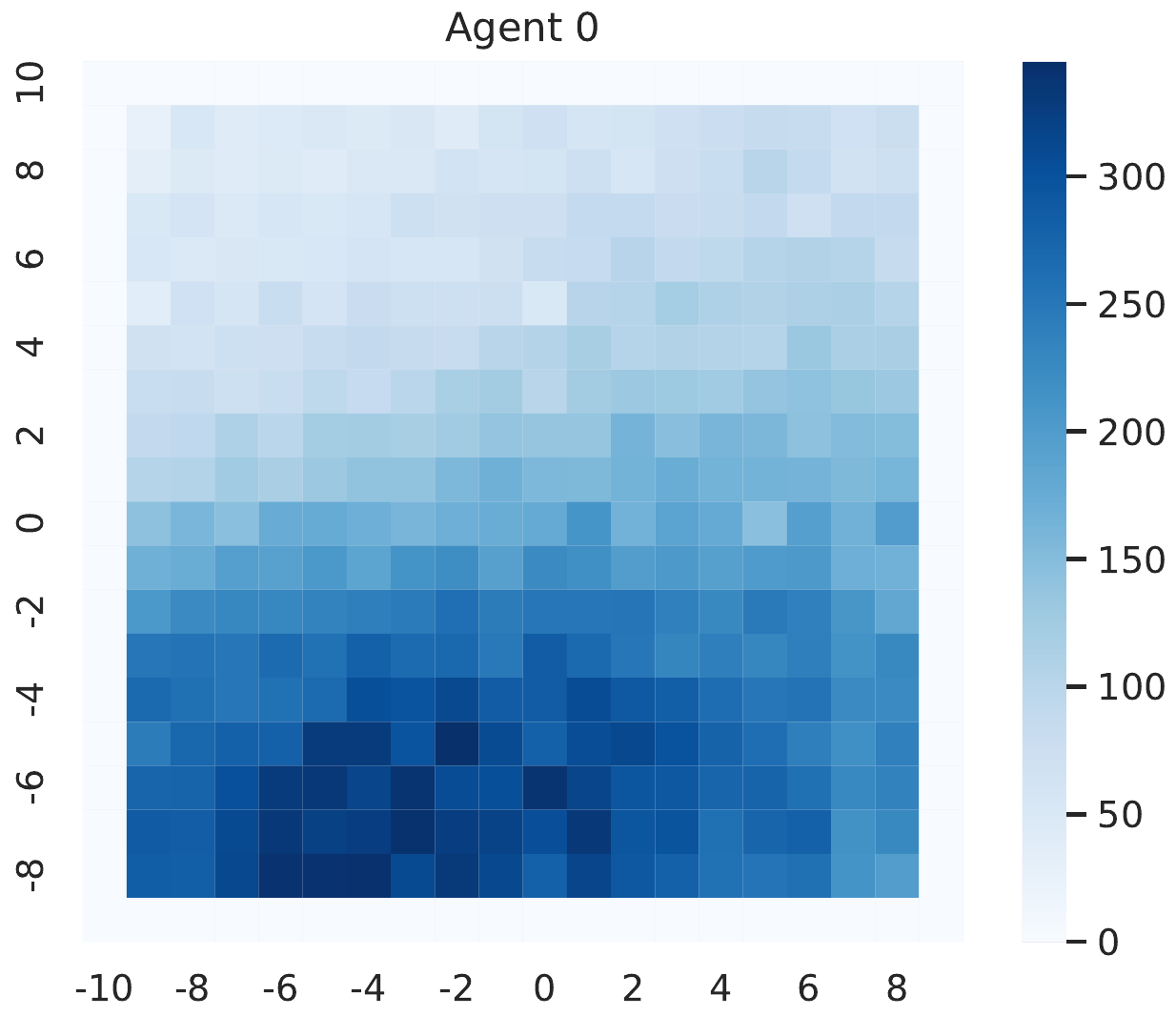}
\end{minipage}\hfil\hfil
\begin{minipage}[t]{0.31\hsize}
    \centering
    \includegraphics[keepaspectratio, width=\linewidth]{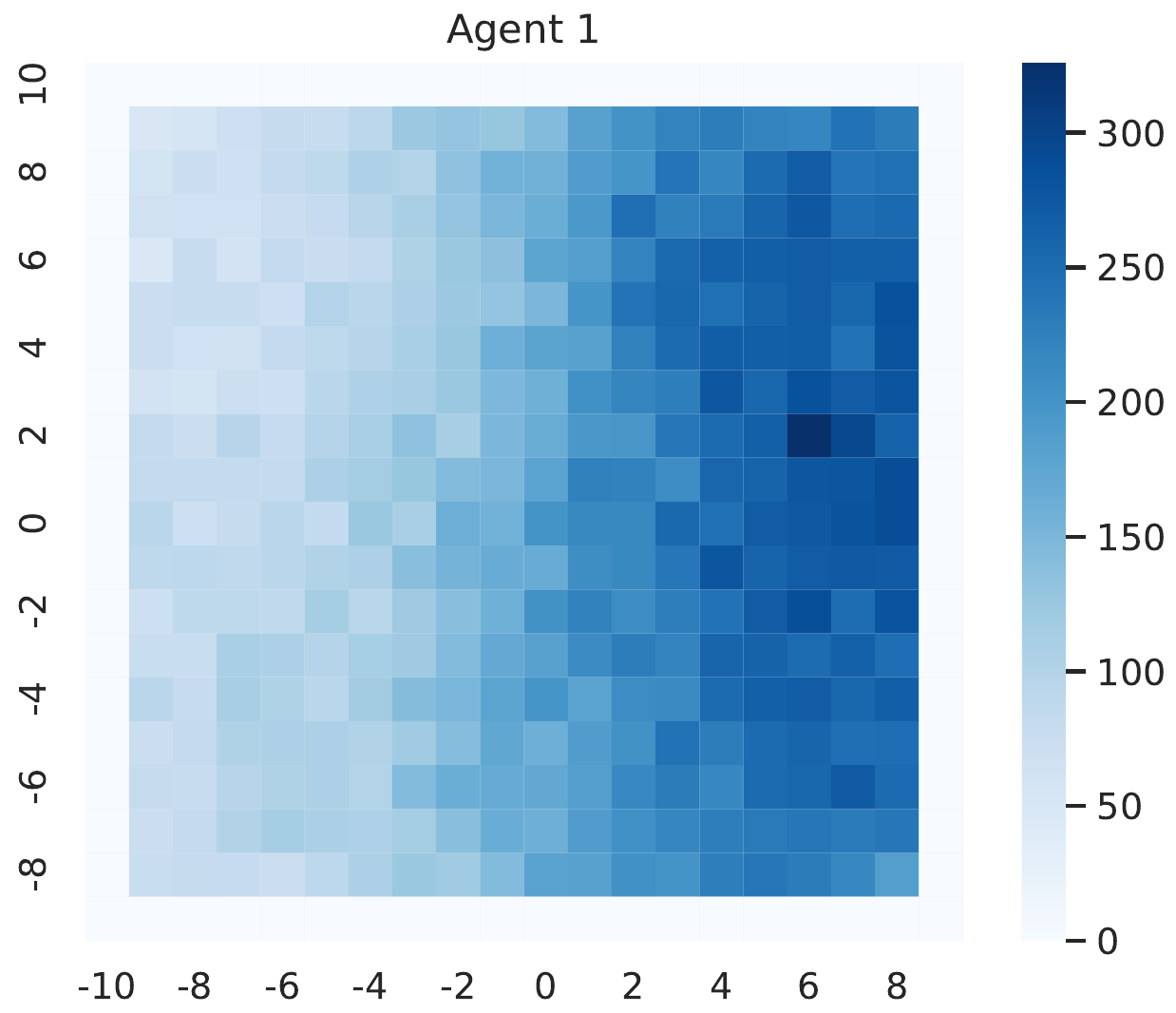}
\end{minipage}\hfil\hfil
\begin{minipage}[t]{0.31\hsize}
    \centering
    \includegraphics[keepaspectratio, width=\linewidth]{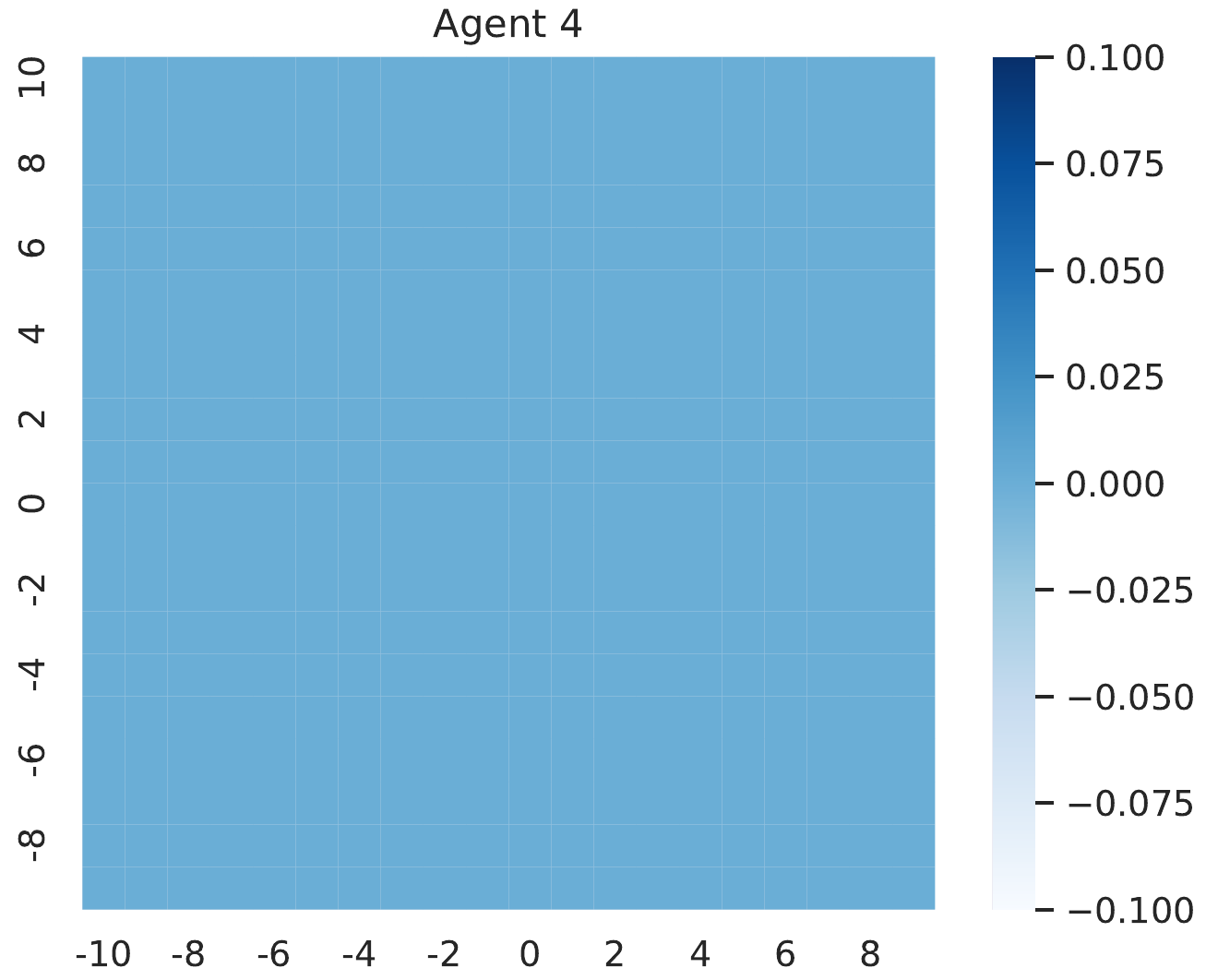}
\end{minipage}\\
\begin{minipage}[t]{0.31\hsize}
    \centering
    \includegraphics[keepaspectratio, width=\linewidth]{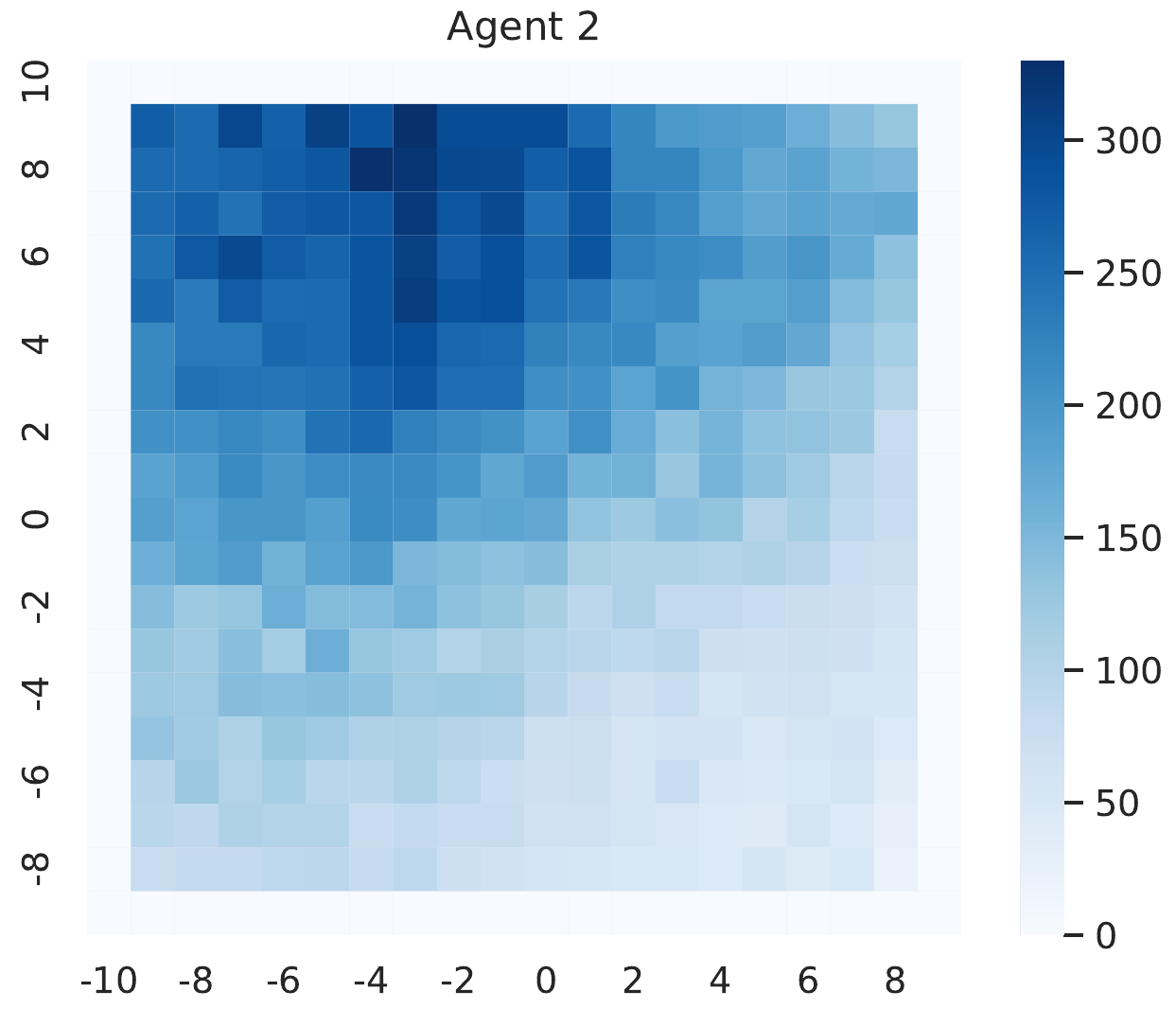}
\end{minipage}\hfil\hfil
\begin{minipage}[t]{0.31\hsize}
    \centering
    \includegraphics[keepaspectratio, width=\linewidth]{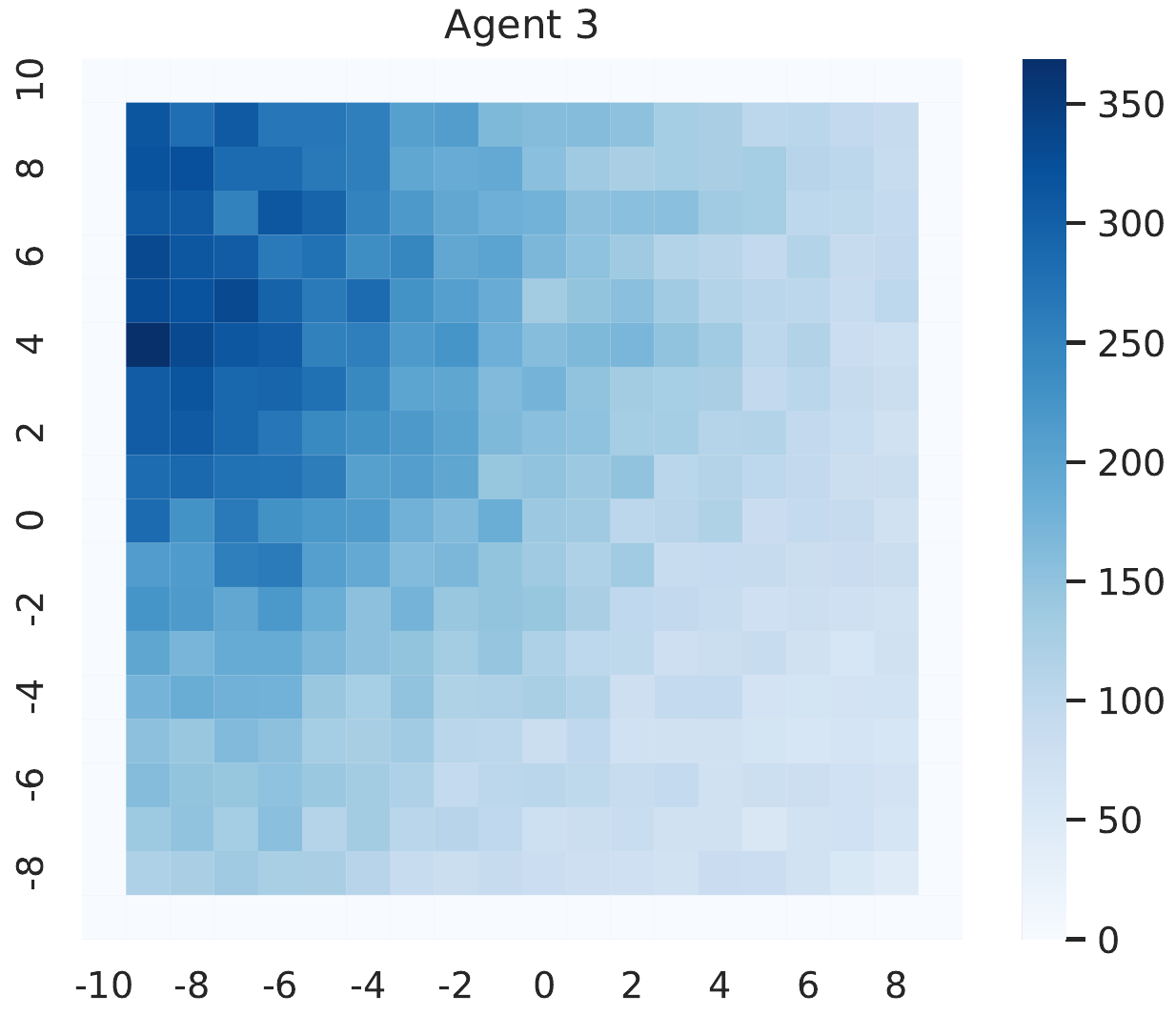}
\end{minipage}\hfil\hfil
\begin{minipage}[t]{0.31\hsize}
    \centering
    \includegraphics[keepaspectratio, width=\linewidth]{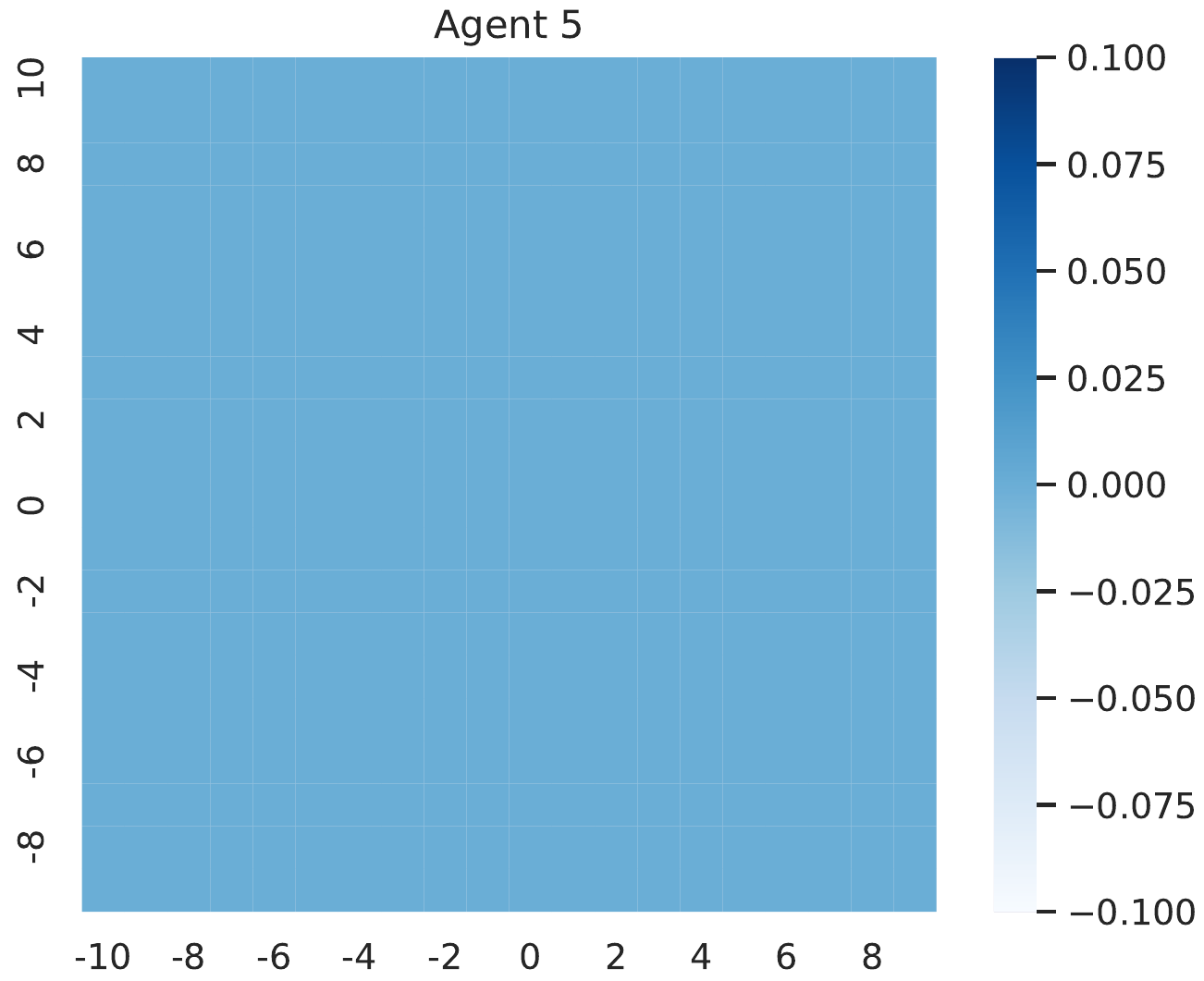}
\end{minipage}
\caption{Object collections of individual agents.}
\label{fig:oc_hm}
\end{figure}

\subsection{Learning with Coordination Noise -- Exp.~2}
Considering the alter-exploration problem, in Exp.~2 we examine the ability of DA3-DQN agents to interpret noise by assessing the appropriateness of agent behaviors in the visible area. The DA3-DQN agents were initially placed on the blue nodes numbered 0, 1, 2, and 3, and two wandering agents were placed on the red nodes numbered 4 and 5 in Env.~2 (see Fig.~\ref{fig:grid-mapb}). The DA3-DQN agents were expected to judge whether to build coordination with other agents within the visible area and/or assess the rationality of their decision-making. If coordination was chosen, the agent does not approach the same object to avoid unnecessary competition; otherwise, irrelevant wandering agents are ignored and the object is approached.
\par

Figure~\ref{fig:coordination_test} demonstrates how one DA3-DQN agent (Agent~0) built coordinated activity in four situations without communication with other agents and exhibited some rationale of agents' decision after training. The sub-figures show each situation by pairing the observations with corresponding heatmaps, with the left pair showing the state at time $t$ and the right pair the state at $t+1$. Note that Agent~0 always pays attention to the upper node of itself (in range of $[0.099, 0.118]$); this phenomenon is not discussed due to the limited number of pages but usually happens when agents build a strategy by focusing on a particular region.
\par

The behavior of Agent~0 was investigated for the first situation where two objects were located at different distances (Situation.~1), as shown in the upper-left heatmap in Fig.~\ref{fig:coordination_test_a}. Its attentional heatmap (the lower-left) indicates that it set attentional weights of $0.125$ and $0.100$ to the closer and further objects, respectively. Then, Agent~0 moved up, and the attentional weight of the closer object increased to $0.183$. Situation.~2 (Fig.~\ref{fig:coordination_test_b}) is similar to Situation.~1 but another agent is located near the closer object. Its attentional heatmap indicates that it assigned attentional weights of $0.124$ $0.100$ and $0.125$ to the closer object, further objects, another cooperative agent (depicted in blue) respectively. Thus, Agent~0 approached the further object to avoid the competition and then assigned a higher attentional weight ($0.116$). This phenomenon convincingly demonstrated the successful building of coordination between DA3-DQN agents.
\par

Situation.~3 (Fig.~\ref{fig:coordination_test_c}) is almost identical to Situation.~2 except that the approaching agent is identified as irrational (shown in red), and Agent~0 assigns the lower attentional weight $0.028$ to this wandering agent. Interestingly, unlike in Situation.~2, Agent~0 moved up for the closer object, whose attentional weight also increased from $0.124$ to $0.182$, because it learned that competition would not occur. Situation.~4 (Fig.~\ref{fig:coordination_test_c}) is also similar to Situation.~3, but the distance between two objects is identical, and there is another cooperative agent near the left object as illustrated in Fig.~\ref{fig:coordination_test_d}. The lower-left heatmap indicates that Agent~0 assigned almost equal attentional weights ($0.109$ and $0.120$) to two objects, a higher weight ($0.087$) to the cooperative agent, and a lower weight ($0.023$) to the irrational agent. Then Agent~0 selected the upper-right object to approach; hence confirming the selective recognition of other agents to determine whether to behave cooperatively by yielding objects.
\par

Finally, we investigated the cooperative behaviors with the trained DA3-DQN agents from a macro perspective. For this purpose, $1,000$ episodes of the same experiment (Exp.~2) were conducted after training, and accumulated heatmaps generated to indicate where each agent collected the objects. These are shown in Fig.~\ref{fig:oc_hm}, where the left four heatmaps are of intelligent agents (0-3) while the right two are of wandering agents (4, 5). Note that the wandering agents can only roam the environment and collect no objects, meaning zero collection is indicated for them. The heatmaps indicate that the DA3-DQN agents roughly divided the square room into four regions, with each agent taking charge of their assigned area. This division of area is due to cooperative behaviors achieved by selecting worthy agents, especially to avoid competition and collision, and ignoring the wandering agents determined by attention mechanism. As a result, the DA3-DQN agents could improve the average episode reward by $3.91\%$.
\par

We believe that such learning mechanisms against noise arising from partly unreliable data and irrelevant agents was unknown in the multi-agent learning context due to the black-box problem. Hence our study further sheds light on explanation for agent decisions during training of deep reinforcement learning.

\section{Conclusion}
We have proposed the DA3-X, which can have any deep reinforcement learning network as a DRL head, for a multi-agent system that can deal with noisy environments as well as improving learning capability. To validate the effectiveness and robustness of our proposed method, we conducted the object collection game with two types of noise: noise in observation and in coordination. The results demonstrated that DA3-X (DA3-IQN and DA3-DQN, in particular) agents outperformed baseline methods in all noisy environments. By analyzing the effect of two types of noise on the agents' decision-making process using the visualized attentional heatmaps from the DA3-DQN, we confirmed that DA3-X agents successfully gave higher weights to the relevant and reliable information in their contaminated observations. We believe this research can be further extended to a more complicated dynamic environment with multiple types of noise.
\par

\section*{Acknowledgment}
This study was partly supported by JSPS KAKENHI Grant Numbers 17KT0044 and 20H04245.


\end{document}